\pdfoutput=1

\documentclass[11pt]{article}

\usepackage[]{acl}

\usepackage{times}
\usepackage{latexsym}
\usepackage{booktabs}
\usepackage{graphicx}
\usepackage{multirow}
\usepackage{colortbl}
\usepackage{pgfplots}
\usepackage{subcaption}

\usepackage[T1]{fontenc}

\usepackage[utf8]{inputenc}

\usepackage{microtype}

\usepackage{inconsolata}

\usepackage{amsmath}

\usepackage{xspace}

\pgfplotsset{compat=1.18} 

\newcommand{\scalemath}[2]{
  \scalebox{#1}{$\displaystyle #2$}
}

\title{On Learning to Summarize with Large Language Models as References}

\author{
 \textbf{Yixin Liu}$^{1}$ 
\quad \textbf{Kejian Shi}$^{1}$ 
\quad \textbf{Katherine S He}$^{1}$ 
\quad \textbf{Longtian Ye}$^{1}$ 
 \\
 \quad \textbf{Alexander R. Fabbri}$^{2}$ 
 \quad \textbf{Pengfei Liu}$^{3}$ 
 \quad \textbf{Dragomir Radev}$^{1}$ 
 \quad \textbf{Arman Cohan}$^{1,4}$ 
  \vspace{4pt}\\
  $^1$Yale University\quad 
  $^2$Salesforce AI \quad 
  $^3$Shanghai Jiao Tong University\quad
  $^4$Allen Institute for AI
  \\
   \texttt{\{yixin.liu,arman.cohan\}@yale.edu}
 }

\begin{document}
\maketitle
\begin{abstract}
Recent studies have found that summaries generated by large language models (LLMs) are favored by human annotators over the original reference summaries in commonly used summarization datasets.
Therefore, we study an LLM-as-reference learning setting for smaller text summarization models to investigate whether their performance can be substantially improved.
To this end, we use LLMs as both oracle summary generators for standard supervised fine-tuning and oracle summary evaluators for efficient contrastive learning that leverages the LLMs' supervision signals.
We conduct comprehensive experiments with source news articles and find that 
(1) summarization models trained under the LLM-as-reference setting achieve significant performance improvement in both LLM and human evaluations;
(2) contrastive learning outperforms standard supervised fine-tuning under both low and high resource settings.
Our experimental results also enable a meta-analysis of LLMs' summary evaluation capacities under a challenging setting, showing that LLMs are not well-aligned with human evaluators.
Particularly, our expert human evaluation reveals remaining nuanced performance gaps between LLMs and our fine-tuned models, which LLMs fail to capture.
Thus, we call for further studies into both the potential and challenges of using LLMs in summarization model development.

\end{abstract}

\section{Introduction}
Recent studies~\cite{Liu2022RevisitingTG, 10.1162/tacl_a_00632, Pu2023SummarizationI} have discovered that large language models (LLMs), like GPT-3.5
~\cite{NEURIPS2022_b1efde53}, can generate summaries that are preferred by human annotators when compared to \textbf{\textit{reference summaries}} from widely used datasets, such as CNN/DailyMail~\cite{Nallapati:16} and XSum~\cite{narayan-etal-2018-dont}, in a \textit{reference-free} human evaluation setting.
This quality issue of existing reference summaries effectively puts an upper bound on the performance of summarization models trained on them, which likely contributes to the performance gap between supervised summarization models, e.g., BART~\citep{lewis-etal-2020-bart}, and LLMs as observed by related work~\cite{Goyal2022NewsSA, liang2023holistic, Liu2022RevisitingTG, 10.1162/tacl_a_00632}.

Therefore, we aim to investigate \textbf{whether smaller summarization models can be substantially improved with better references}.
To this end, we study an \textit{\textbf{LLM-as-reference}} distillation setting, where the LLMs are considered the reference or the gold-standard oracle for the summarization task.
Specifically, we employ LLMs in the training of smaller text summarization models in two manners:
(1) LLMs as the gold summary generator, where the model is trained with the LLM summary as the reference under the standard supervised fine-tuning;
(2) LLMs as the gold summary evaluator, where LLM-based automatic evaluation methods~\cite{Fu2023GPTScoreEA, liu-etal-2023-g} are used as supervision signals for training techniques such as contrastive learning~\citep{liu-etal-2022-brio, zhao2023calibrating} and reinforcement learning~\citep{paulus2018a, NEURIPS2020_1f89885d}.

Using the source articles in the CNN/DailyMail dataset, we conduct comprehensive experiments for this LLM-as-reference setting, across proprietary and open-source LLMs under low and high resource conditions.
The experimental results demonstrate that (1) LLM-generated summaries are better references for the smaller models than the original reference summaries, and (2) contrastive learning with LLMs as evaluators outperforms standard supervised fine-tuning.
In particular, our best-performing fine-tuned BART checkpoint can outperform GPT-3.5
under GPT-4's evaluation~\cite{OpenAI2023GPT4TR}.
Meanwhile, under expert human evaluation, it can achieve similar or superior overall performance to GPT-3.5 in 50\% of cases, while the original fine-tuned BART has a comparable rate of success in only 4\% of cases.

To have a more comprehensive understanding of this LLM-as-reference setting, we then conduct a meta-analysis of the LLM-based evaluation methods by assessing their alignment level to the human evaluation.
While these LLM-based methods achieve strong performance in existing meta-evaluation datasets consisting of summaries generated by supervised summarization systems~\cite{Fu2023GPTScoreEA, liu-etal-2023-g}, we find that they do not correlate well with human evaluation when comparing close-performing systems in our setting.
Particularly, although our fine-tuned model achieves better performance than LLMs under LLM-based evaluation, human evaluation reveals remaining nuanced performance gaps between our model and LLMs.

Our main contributions are two-fold: 
(1) We empirically demonstrate that the performance of smaller models can be substantially improved when trained using better references (LLMs) and learning methods (contrastive learning).
(2) We perform a meta-analysis of LLM-based evaluation under a challenging scenario enabled by our task setting where LLMs need to compare summarization systems with close performance, which indicates that LLMs fail to align with human evaluation and capture nuanced performance differences.\footnote{We release the scripts, training data, and model outputs at \url{https://github.com/yixinL7/SumLLM}.}

\section{Methods}
\label{sec:methods}
\subsection{Preliminary}

A neural abstractive summarization model $g$ aims to generate a text sequence $S$ that summarizes the information of a source document $D$: $S \leftarrow g(D)$.
When $g$ is an \textit{auto-regressive} text generation model, it factorizes the probability of a candidate summary $S$ given the source document $D$ as 
\begin{equation}
\label{eq:auto}
\scalemath{0.9}{p_g(S|D) = \prod_{i=1}^{l_S} p_g(s_i|S_{<i}, D),}
\end{equation}
where $s_i$ is the $i$-th token in $S$ and $s_0$ is a special begin-of-sequence (BOS) token, $S_{<i}$ is the prefix-string of $S$ before $s_i$, $l_S$ is the length of $S$ (without the BOS token), and $p_g$ is a probability distribution parameterized by the summarization model $g$.

The standard training algorithm for $g$ is Maximum Likelihood Estimation (MLE) with a single reference (gold standard) summary $S^*$.
With Eq.~\ref{eq:auto}, the MLE optimization on this example is equivalent to minimizing the following cross-entropy loss:
\begin{equation}
\label{eq:xent}
\scalemath{0.9}{\mathcal{L}_{xent}(\theta) =  - \log p_{g} (S^* |D; \theta),}
\end{equation}
where $\theta$ are the learnable parameters of $g$.

\subsection{Large Language Models as References}

Similar to Eq.~\ref{eq:auto}, an auto-regressive LLM $h$ defines a target distribution for text summarization:
\begin{equation}
\label{eq:llm}
\scalemath{0.9}{ p_h(S|D) = \prod_{i=1}^{l_S} p_h(s_i|S_{<i}, D),}
\end{equation}
which is different from the point-mass distribution defined by a single reference summary (Eq.~\ref
{eq:xent}).
Consequently, the cross-entropy loss becomes
\begin{equation}
\scalemath{0.9}{\mathcal{L}_{xent}(\theta; h) = - \sum_{S \in \mathcal{S}} p_h (S|D) \log p_{g} (S |D; \theta),}
\label{eq:xent-llm}
\end{equation}
where $\mathcal{S}$ is the set of possible outputs (candidate summaries).
This setting is coined \textit{\textbf{sequence-level knowledge distillation}} by \citet{kim-rush-2016-sequence}.
In practice, computing Eq.~\ref{eq:xent-llm} is intractable because $\mathcal{S}$ is infinite.
Thus, we explore various approaches to approximate this learning objective.

\subsubsection{LLMs as Gold Summary Generators}

\paragraph{MLE Fine-tuning} Our baseline method treats the greedy decoding results of the LLM $h$ as the reference summaries and optimizes the summarization model $g$ using MLE.
The loss function becomes
\begin{equation}
\scalemath{0.9}{\mathcal{\hat{L}}_{xent}(\theta; h) = - \log p_{g} (\hat{S} |D; \theta),}
\label{eq:xent-llm-quasi}
\end{equation}
where $\hat{S}$ is the greedy decoding result of $h$:
\begin{equation}
\label{eq:greedy}
\scalemath{0.9}{\hat{s}_i = \arg\max_s p_h(s|\hat{S}_{<i}, D),}
\end{equation}
where $s$ denotes a token in the vocabulary.

\paragraph{Contrastive Learning} To improve the performance beyond MLE, we adopt a contrastive learning method, BRIO~\citep{liu-etal-2022-brio}, for \textit{reference-based} model training, which sets the following training objective: given two candidate summaries $S_i$, $S_j$, if $S_i$ is better than $S_j$, the summarization model $g$ should assign $S_i$ a higher probability (Eq.~\ref{eq:auto}).
In more detail, this loss is defined with a set of candidate summaries $\mathcal{S}_c$, which is \textit{descendingly sorted} by their \textit{\textbf{similarity with the reference summary}}, as measured by an automatic metric such as ROUGE~\cite{lin-2004-rouge}.
The summarization model $g$ is tasked with assigning a probability that is at least twice\footnote{We found that in practice the model training is insensitive to this value so we set it to a constant value for simplicity.} as large to a better candidate:
\begin{equation}
\scalemath{0.9}{\frac{p_g(S_i|D)}{p_g(S_j|D)} > 2(j - i), \forall i, j, i < j,}
\end{equation}
which corresponds to the following margin loss:

\scalebox{0.9}{
\begin{minipage}{\linewidth}
\begin{align}
\label{eq:ctr-1}
 \mathcal{L}_{ctr}(\theta) &= \sum_{S_i, S_j \in \mathcal{S}_c, i < j} \max(0,  \log p_g(S_j|D; \theta) \nonumber \\
    & - \log p_g(S_i|D; \theta) + \log 2(j - i)).
\end{align}
\vspace{-10pt}
\end{minipage}
}
%

Following \citet{liu-etal-2022-brio}, we combine the cross-entropy loss (Eq.~\ref{eq:xent-llm-quasi}) with the contrastive loss as a multi-task loss:
\begin{equation}
\label{eq:multi}
\scalemath{0.9}{\mathcal{L}_{mul}(\theta) = \mathcal{\hat{L}}_{xent}(\theta; h) +  \alpha \mathcal{L}_{ctr}(\theta),}
\end{equation}
where $\alpha$ is the weight of the contrastive loss.

\subsubsection{LLMs as Gold Summary Evaluators}
\label{subsec:llm-as-evaluator}
Apart from the reference summaries, LLMs can also provide \textit{reference-free} supervision signals for model training since they can be used to evaluate the quality of any candidate summary.
As these LLM-based evaluation methods have shown superior performance than traditional metrics such as ROUGE~\cite{Fu2023GPTScoreEA, liu-etal-2023-g}, we hypothesize that they can provide more accurate supervision that enables efficient training.
Consequently, we expand the contrastive learning approach (Eq.~\ref{eq:ctr-1}) by using LLM-based evaluation to provide the gold ranking of the candidate summaries.
We focus on two recent LLM-based evaluation methods: GPTScore~\citep{Fu2023GPTScoreEA} and an extended version of G-Eval~\cite{liu-etal-2023-g}, which we coin GPTRank.

\paragraph{GPTScore for Summary Quality Evaluation}
The contrastive learning objective (Eq.~\ref{eq:ctr-1}) requires access to ground-truth candidate summary quality scores from the reference LLM.
Therefore, we first adopt GPTScore~\citep{Fu2023GPTScoreEA} for the summary quality evaluation.
Specifically, GPTScore interprets the length-normalized conditional log-probability of a candidate summary predicted by the reference LLM $h$ as its quality score, i.e., 
\begin{equation}
\label{eq:log-normalized-llm}
\scalemath{0.9}{\bar{p}_h(S|D) = \frac{\sum_{i=1}^{l_S} \log p_h(s_i|S_{<i}, D)}{l_S}.}
\end{equation}
Consequently, the set of candidate summaries $\mathcal{S}_c$ used in Eq.~\ref{eq:ctr-1} is sorted based on the (normalized) target distribution (Eq.~\ref{eq:llm}), such that for any $S_i, S_j \in \mathcal{S}_c, i < j$, $\bar{p}_h(S_i|D) > \bar{p}_h(S_j|D)$.

\paragraph{GPTRank for Summary Quality Evaluation}

Instead of leveraging the LLM predicted probability, recent work, e.g., G-Eval~\citep{liu-etal-2023-g}, formulates the automatic evaluation as a text completion or infilling task for the LLMs, requiring them to provide a numerical quality score for an evaluation task.
We extend this evaluation method, which we coin \textbf{GPTRank}, by requiring the LLM to provide a \textit{\textbf{quality ranking}} to a list of different candidate summaries for the same source article.
Moreover, since recent work~\cite{liu-etal-2023-g} has found that language models can benefit from a self-explaining stage for an evaluation task, we prompt the LLM to first generate an \textit{\textbf{explanation}} before providing the actual ranking.
The ranking is then used in contrastive learning (Eq.~\ref{eq:ctr-1}). 

\section{Learning with LLMs as References}
\label{sec:exp}

We conduct experiments with both proprietary and open-source LLMs in the LLM-as-reference learning setting of smaller summarization models across low and high resource conditions and compare different training methods.

\subsection{Learning under Low Resource Settings}
Proprietary LLMs, such as GPT-4, can be more capable than open-source LLMs but are less cost-efficient.
Therefore, we focus on a cost-effective low-resource setting using contrastive learning where the LLMs are used as both summary generators and evaluators.

\subsubsection{Experimental Setting}
\label{subsec:setting}

\paragraph{Data Source} 
We use mainly source articles from the CNN/DailyMail (CNNDM) dataset for our experiments, and 100 test examples are sampled for LLM-based and human evaluation.
The LLMs are prompted to generate three-sentence summaries to approximate the original summary style
with a 0 sampling temperate to approximate the greedy decoding process (Eq.~\ref{eq:greedy}).\footnote{Further information regarding the prompts and the process of generating LLM summaries can be found in Appendix \ref{appendix:llm-gen}.}

\paragraph{Training Details}
We choose BART as the smaller summarization model for fine-tuning because it is widely used and is relatively small with around 350 million parameters.\footnote{\url{https://huggingface.co/facebook/bart-large}}
The fine-tuning process involves an MLE warm-up stage with around 10K GPT-3.5 summaries\footnote{We used the checkpoint \texttt{gpt-3.5-turbo-0301} at \url{https://platform.openai.com/docs/models/gpt-3-5}.} and further MLE training or contrastive learning using a reference LLM with around 100 - 1000 training examples.
During the experiments, we compare the model performance trained under MLE and contrastive learning, while making sure that the LLM API cost is similar under the two settings for fair comparison.

For contrastive learning, 8 candidate summaries are used on each training example, generated by MLE-finetuned model checkpoints.
Further experimental details can be found in \ref{appendix:experiment-details}.

\paragraph{Automatic Evaluation} %
For \textit{reference-based} evaluation, we report the ROUGE-1/2 F1 scores between the system outputs and the reference summaries generated by the reference LLM.
For \textit{reference-free} evaluation, we use either GPTScore \cite{Fu2023GPTScoreEA} or GPTRank (\S\ref{subsec:llm-as-evaluator}).
In particular, for GPTScore we report both the un-normalized and normalized sum of log-probability.

\paragraph{Baseline Models}
The following model's performance is compared: (1) GPT3D3, (2) the BART checkpoint fine-tuned on the original CNNDM dataset, (3) GPT3D2 (OpenAI's \texttt{text-davinci-002}), (4) a 7B Alpaca checkpoint,\footnote{\url{https://github.com/tatsu-lab/stanford_alpaca}} (5) GPT-3.5 (OpenAI's \texttt{gpt-3.5-turbo-0301}).

\subsubsection{Learning with GPTScore}
\label{subsec:gptscore}

\begin{table}[t!]
\small
\centering
\addtolength{\tabcolsep}{-1.5pt} 
\begin{tabular}{lccccc}
\toprule
\textbf{System} & \textbf{LP} & \textbf{GS} & \textbf{R1}  & \textbf{R2} & \textbf{Len.}\\
\midrule
GPT3D3       & -22.62 & -0.271 &   100.0 &   100.0 &     85.4 \\
\midrule
 BART         & -59.55 & -0.789 &    46.85 &    24.38 &     79.0 \\
 GPT3D2       & -41.21 & -0.547 &    55.40 &    33.72 &     78.7 \\
 Alpaca       & -44.82 & -0.567 &    51.53 &    30.18 &     81.8 \\
 GPT-3.5      & -45.12 & -0.498 &    58.14 &    37.46 &     92.0 \\
\midrule
BART.GPT3D3  & -36.13 & -0.420 &    \textbf{59.50} &    \textbf{40.70} &     85.6 \\
 BRIO.GPT3D3  & \textbf{-26.20} & \textbf{-0.318} &    56.21 &    36.47 &     83.7 \\
\bottomrule
\end{tabular}
\addtolength{\tabcolsep}{+1.5pt} 
\caption{Results with GPTScore. 
\textbf{LP} is the log-probability predicted by \textbf{GPT3D3}.
\textbf{GS} is the GPTScore based on GPT3D3.
\textbf{R1} and \textbf{R2} are the ROUGE1/2 F1 scores respectively. 
\textbf{Len.} is the average summary length. 
\textbf{BART.GPT3D3} is fine-tuned with MLE training while \textbf{BRIO.GPT3D3} is fine-tuned with contrastive learning.
}
\label{tab:main-result-gptscore} 
\end{table}

For GPTScore, the reference LLM we choose is OpenAI's \texttt{text-davinci-003} (\textbf{GPT3D3}), since its API provides access to the predicted log-probability.\footnote{We note that the more recent OpenAI models, such as GPT-4, do not provide log-probability of the input tokens.}
We report the model performance in Table~\ref{tab:main-result-gptscore}, with the following observations:

\noindent (1) Compared with the original BART checkpoint, MLE training on reference summaries from LLMs can effectively improve the model performance as measured by either GPTScore or ROUGE.

\noindent (2) The model trained with contrastive learning (BRIO.GPT3D3) can achieve significantly better GPTScore than the model fine-tuned with MLE training (BART.GPT3D3), demonstrating the effectiveness of contrastive learning for approximating the target distribution of the reference LLM.

\noindent (3) BRIO.GPT3D3 can already achieve a similar GPTScore as the reference LLM (GPT3D3) itself while only being trained on 100 examples with contrastive learning, showing a promising path to further close the performance gap.

\subsubsection{Learning with GPTRank}
\label{subsec:exp-gptrank}

\begin{figure*}[t!]
    \centering
    \begin{subfigure}[b]{0.49\linewidth}
        \centering
        \includegraphics[width=\linewidth]{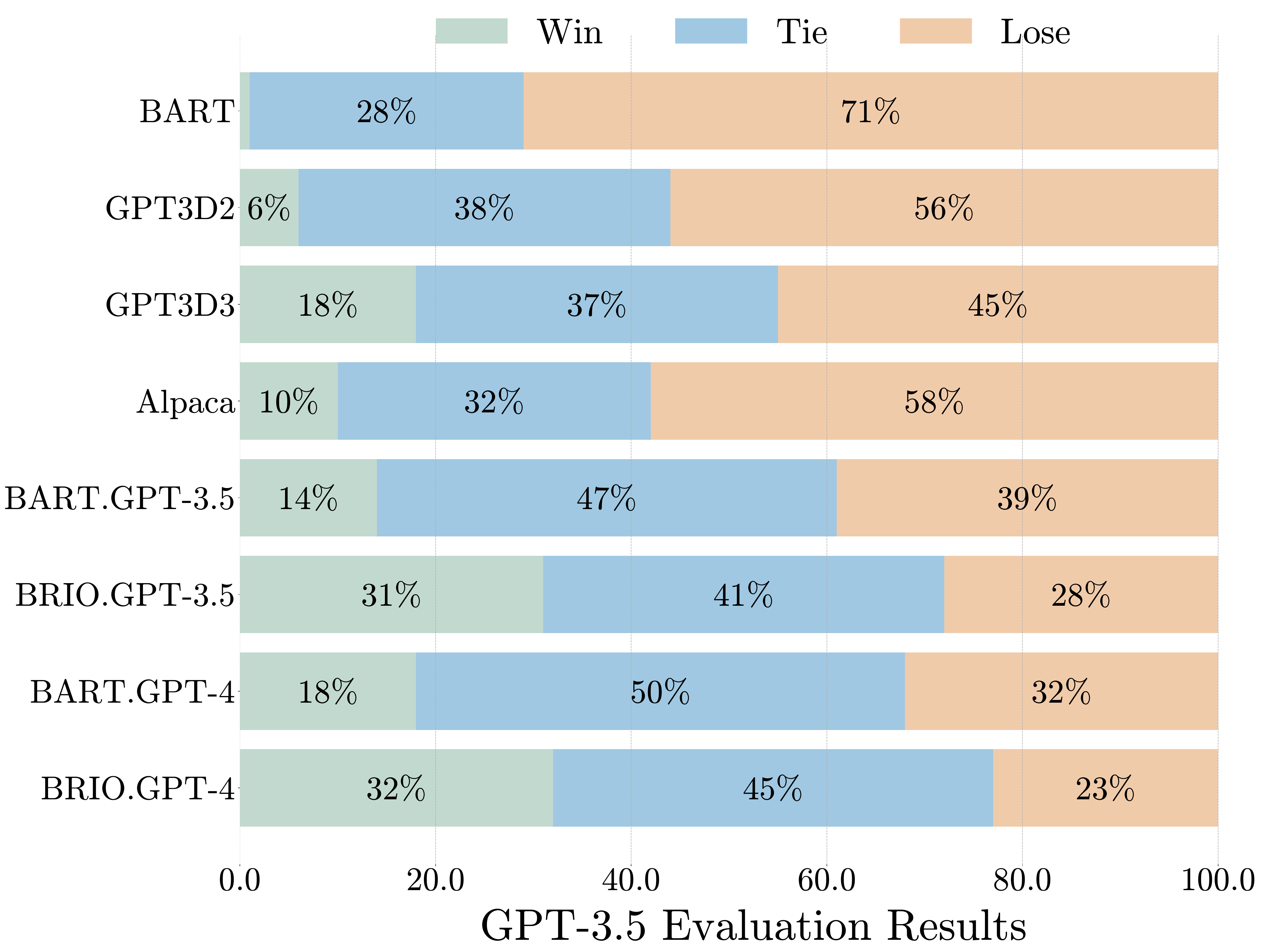}
        \label{fig:gpt3.5compare}
    \end{subfigure}
    \hfill
    \begin{subfigure}[b]{0.49\linewidth}
        \centering
        \includegraphics[width=\linewidth]{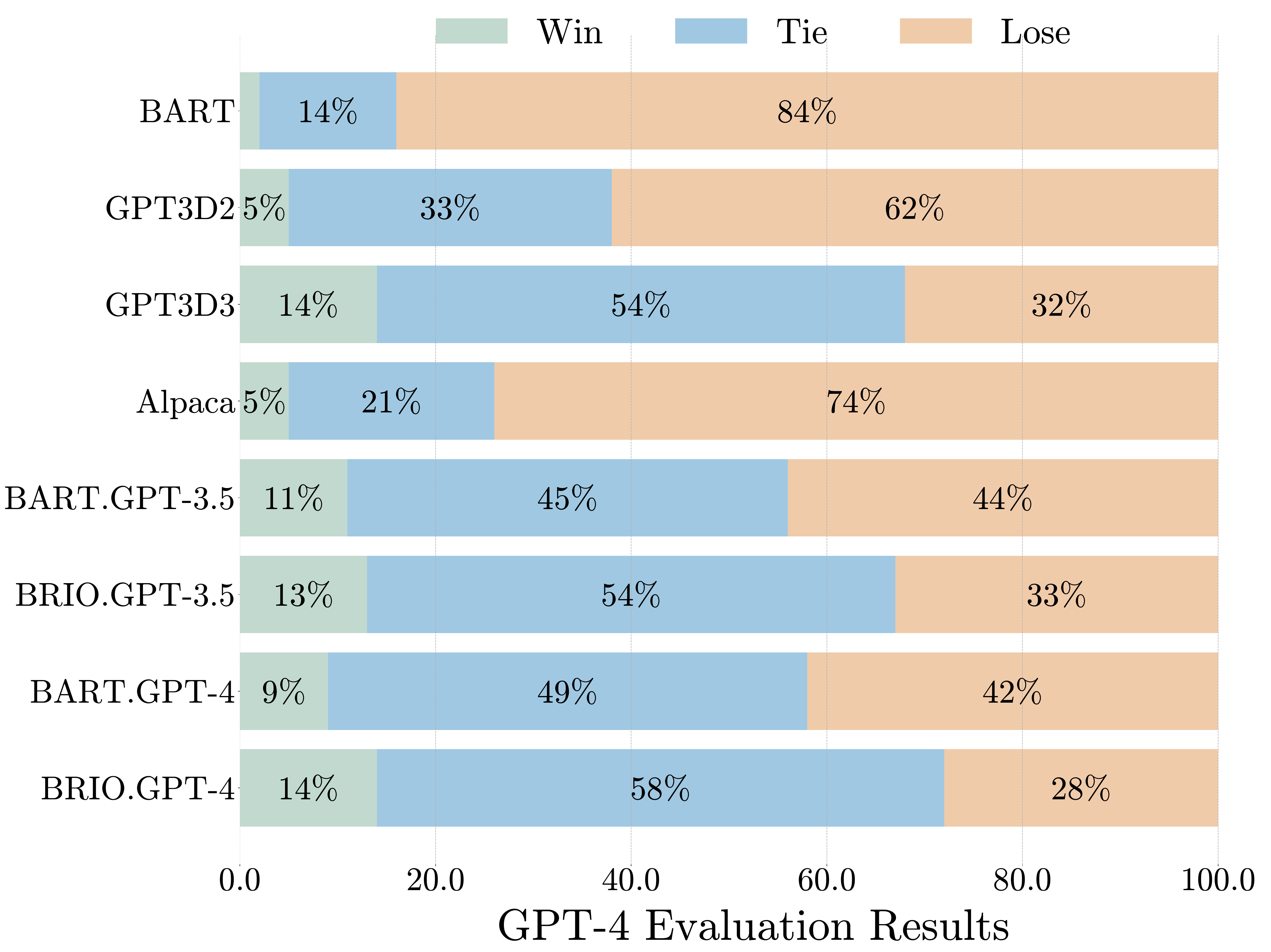}
        \label{fig:gpt4compare}
    \end{subfigure}
    \vspace{-5mm}
    \caption{Pairwise comparison (GPTRank) results of different models \textit{against GPT-3.5} under GPT-3.5's evaluation (left) and GPT-4's evaluation (right).
    BART.GPT-3.5 and BART.GPT-4 are fine-tuned with MLE training and GPT-3.5/GPT-4 as the reference, BRIO.GPT-3.5 and BRIO.GPT-4 are fine-tuned with contrastive learning.}
    \label{fig:gptcompare}
\end{figure*}

We now conduct experiments using GPTRank for model training and evaluation.
The reference LLMs we choose are GPT-3.5 and GPT-4~\cite{OpenAI2023GPT4TR} since they have shown state-of-the-art performance on summarization evaluation~\cite{liu-etal-2023-g}.\footnote{We use the \texttt{GPT-4-0314} version: \url{https://platform.openai.com/docs/models/GPT-4}.}
To enable a more accurate evaluation, we choose GPT-3.5 as the baseline model and use the LLMs to conduct a \textit{pairwise} comparison between different systems and GPT-3.5.
To reduce the positional bias in LLM evaluation results as noted by \citet{wang2023large}, we evaluate each summary pair in \textit{both} directions and take the average of results.
In addition, we allow the LLMs to predict a tie between two summaries.\footnote{The prompt templates are shown in Appendix \ref{appendix:gptrank}.} 

In Figure~\ref{fig:gptcompare}, we report the pairwise comparison results of different models against GPT-3.5 under both GPT-3.5 and GPT-4's evaluation.
We note:

\noindent (1) As in \S\ref{subsec:gptscore}, using better references and contrastive learning helps the model to achieve better LLM-based evaluation results.

\noindent (2) Interestingly, GPT-3.5 prefers both BRIO.GPT-3.5  and BRIO.GPT-4 over its own outputs in the pairwise comparison, suggesting that contrastive learning can efficiently optimize the summarization model for a specific evaluation metric.

\noindent (3) LLM-based evaluation results vary across different LLMs.
For example, while GPT-3.5 prefers BRIO.GPT-4 over itself, GPT-4 prefers GPT-3.5.

\noindent (4) BRIO.GPT-3.5 can outperform BART.GPT-4 despite the fact that BRIO.GPT-3.5 is trained with a reference LLM that is supposedly weaker, indicating the advantage of contrastive learning.

The reference-based evaluation results can be found in Table~\ref{tab:gptrank-ref}.

\begin{table}
    \centering
    \small
\addtolength{\tabcolsep}{-2.0pt} 
\begin{tabular}{lccccc}
\toprule
 \multirow{2}{*}[-2pt]{\textbf{System}}         &     \multicolumn{2}{c}{\textbf{GPT-3.5}} & \multicolumn{2}{c}{\textbf{GPT-4}} & \multirow{2}{*}[-2pt]{\textbf{Length}}\\
 \cmidrule{2-3} \cmidrule{4-5}  
 & \textbf{R1} & \textbf{R2} & \textbf{R1} & \textbf{R2} & \\
\midrule
  GPT-3.5      & 63.43   & 44.09   & 100.0  & 100.0  & 92.0   \\
 GPT-4        & 100.0  & 100.0  & 63.43   & 44.09   & 90.0   \\
\midrule
 BART         & 50.83   & 29.47   & 50.54   & 29.31   & 79.0   \\
 GPT3D2       & 55.17   & 33.23   & 55.34   & 33.31   & 78.7   \\
 GPT3D3       & 56.12   & 34.72   & 58.14   & 37.46   & 85.4   \\
 Alpaca       & 54.77   & 33.23   & 53.41   & 31.48   & 81.8   \\
\midrule
 BART.ChatGPT & 59.52   & 40.45   & 62.04   & \textbf{43.76}   & 94.1   \\
 BRIO.ChatGPT & 57.56   & 35.74   & 61.40   & 40.74   & 93.1   \\
 BART.GPT4    & \textbf{63.22}   & \textbf{44.70}   & 62.08   & 43.55   & 91.8   \\
 BRIO.GPT4    & 58.65   & 37.57   & \textbf{62.79}   & 43.65   & 92.8   \\
\bottomrule
\end{tabular}
\addtolength{\tabcolsep}{2.0pt} 
    \caption{Reference-based evaluation results of GPTRank-based training. GPT-3.5 and GPT-4's summaries are used as the references. \textbf{R1} and \textbf{R2} are the ROUGE1/2 F1 scores respectively. 
\textbf{Len.} is the average summary length. }
    \label{tab:gptrank-ref}
\end{table}

\subsubsection{Comparative Study}
\label{subsec:comparative-study}

We investigate the generalization ability of our training method regarding the choice of the backbone model and the data format.

\begin{figure}[t!]
    \centering
         \includegraphics[width=1.0\linewidth]{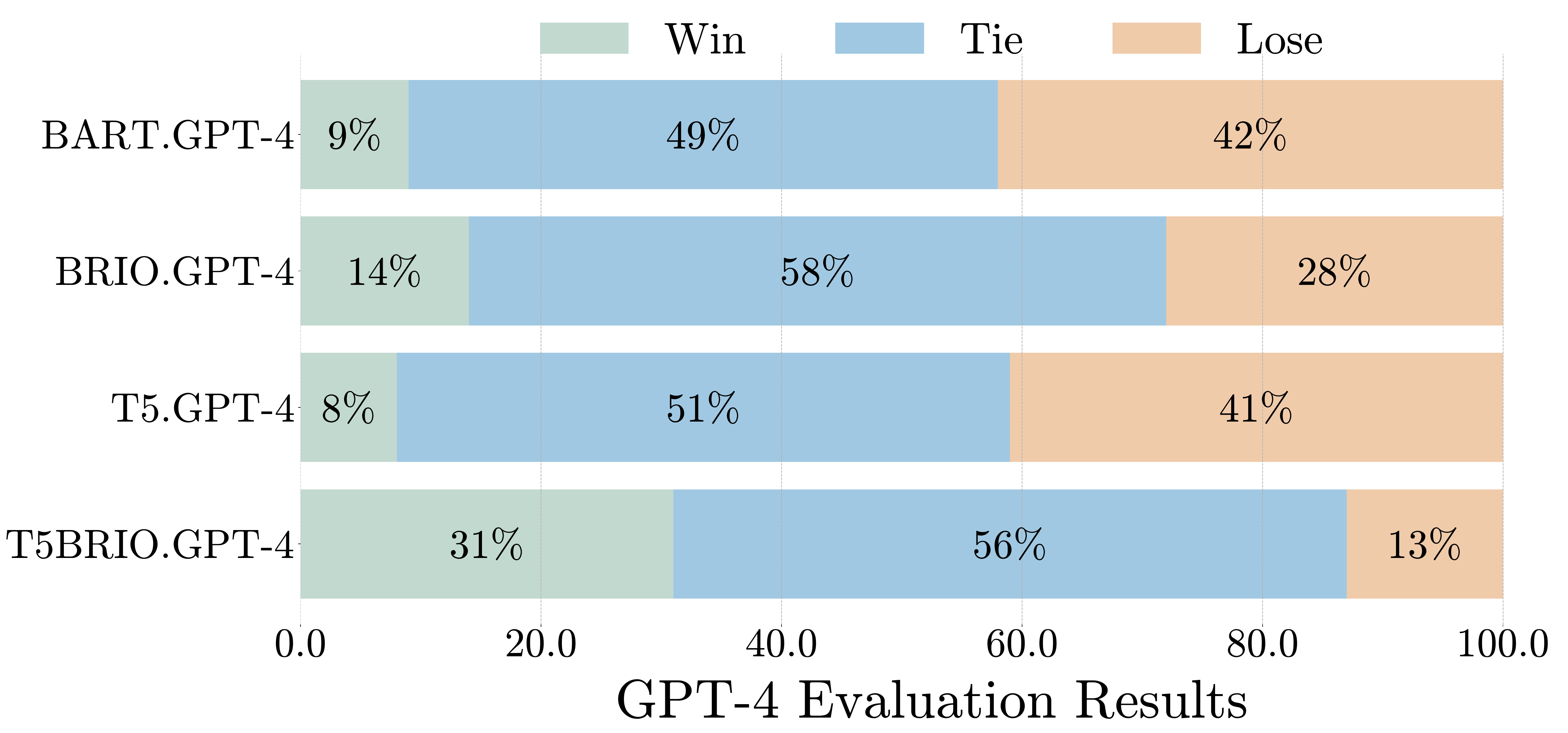}
 \caption{Results of T5 and BART models compared against GPT-3.5 under GPT-4's evaluation.
   BART.GPT-4 and T5.GPT-4 are MLE fine-tuned, BRIO.GPT-4 and T5BRIO.GPT-4 are fine-tuned with contrastive learning.}
 \label{fig:gptcompare-t5}
\end{figure}

\noindent\textbf{Experiments with FLAN-T5}
We repeat the experiment in \S\ref{subsec:exp-gptrank} but use a three billion FLAN-T5~\cite{Chung2022ScalingIL} model\footnote{\url{https://huggingface.co/google/flan-t5-xl}} as the backbone model.
Results in Figure~\ref{fig:gptcompare-t5} suggest that the choice of training algorithms can be more important than the model size for model performance, as BRIO.GPT-4 can outperform T5.GPT-4.
The FLAN-T5 checkpoint trained with contrastive learning, T5BRIO.GPT-4, achieves a strong performance.
However, we note that its summaries are significantly longer than those of other systems, which makes the result more difficult to interpret as recent work has found a strong correlation between the summary rating and length in both human and LLM-based summarization evaluation~\cite{Liu2022RevisitingTG, rajani2023llm_labels}.
Further discussion is in Appendix \ref{subsec:t5-analysis}.

\begin{figure}[t!]
    \centering
         \includegraphics[width=1.0\linewidth]{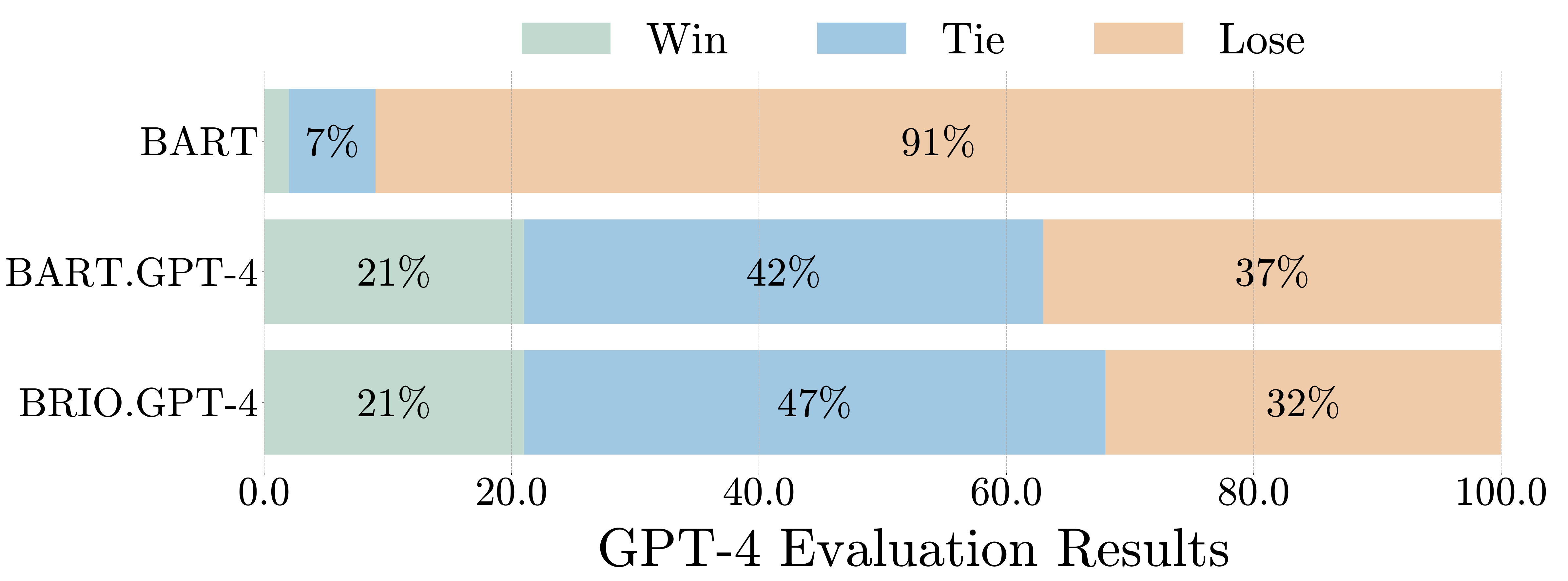}
 \caption{
 Results on XSum dataset. Different models are compared against GPT-3.5 under GPT-4's evaluation.
  BART.GPT-4 is fine-tuned with MLE training while BRIO.GPT-4 is fine-tuned with contrastive learning.}
 \label{fig:gptcompare-xsum}
\end{figure}

\noindent\textbf{Experiments on XSum}
We conduct experiments on XSum~\cite{narayan-etal-2018-dont}, another commonly used dataset.
We follow the original XSum data format by having the models generate one-sentence summaries. 
The experimental settings are similar to those in \S\ref{subsec:setting} \& \S\ref{subsec:exp-gptrank} and more details are in Appendix \ref{subsec:xsum-appendix}.
The results in Figure~\ref{fig:gptcompare-xsum} show a similar trend in that training with better references helps to improve model performance. 

\subsection{Learning under High Resource Settings}
\label{subsec:llama2}

\begin{figure}[t!]
    \centering
         \includegraphics[width=1.0\linewidth]{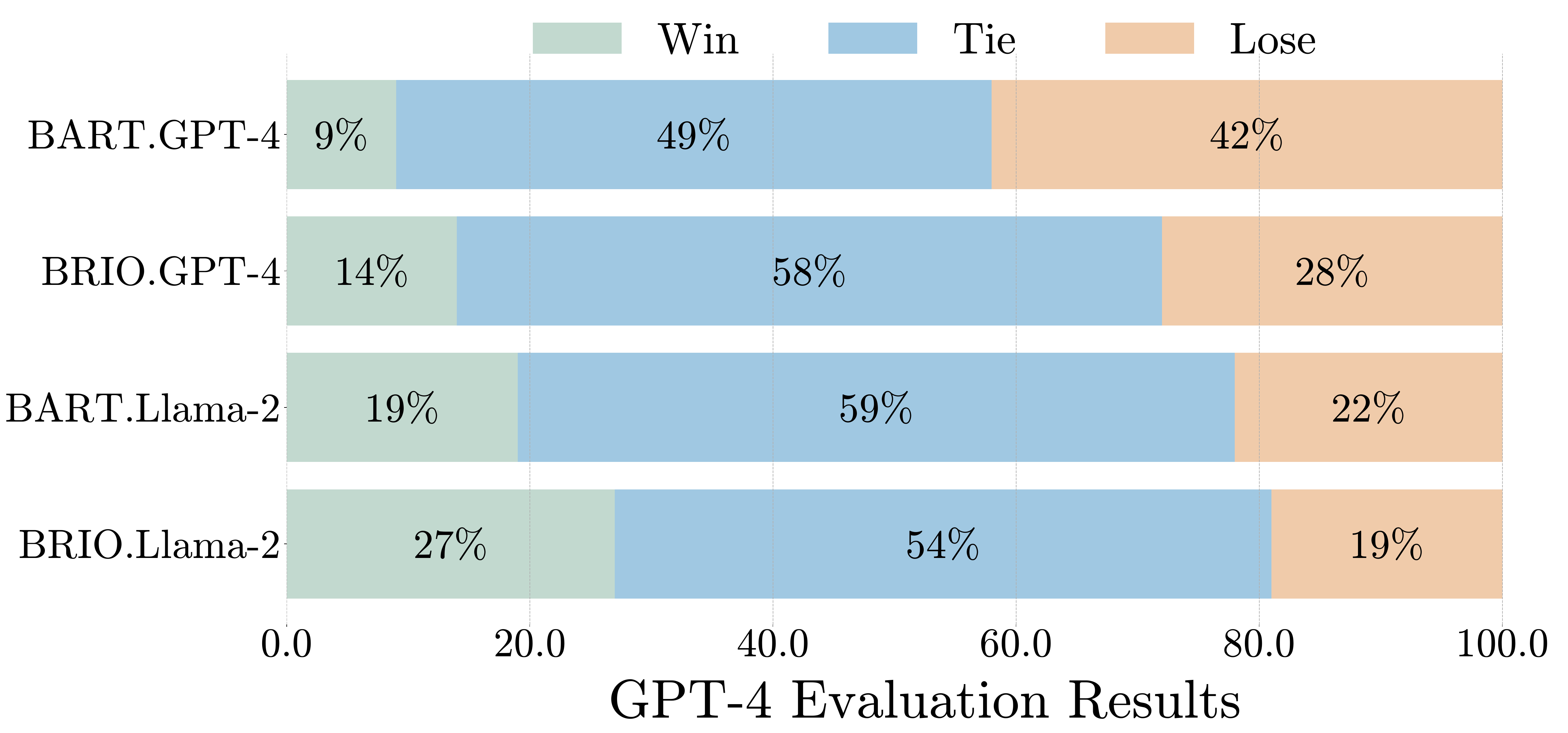}
 \caption{
 Model performance under low and high resource settings. Models are compared against GPT-3.5 under GPT-4's evaluation. The models trained with Llama-2 are under high resource settings and the models trained with GPT-4 are under low resource settings.}
 \label{fig:gptcompare-llama}
\end{figure}

Open-source LLMs provide easier access than proprietary LLMs, however, their performance can be worse, especially for complicated evaluation tasks~\cite{liu2023benchmarking}.
Therefore, we now investigate a high resource setting, where open-source LLMs are used as summary \textit{generators} only to obtain a large number of reference summaries.

\subsubsection{Experimental Setting}

We use the Llama-2 7B Chat model~\cite{touvron2023llama} to generate around 280K reference summaries for the model training.
We fine-tune the BART model using both MLE training and \textit{reference-based} contrastive learning, where the candidate summaries in contrastive learning are ranked by a reference-based automatic evaluation metric.
For contrastive learning, the candidate summaries are ranked based on their content similarity to the Llama-2 summaries, rather than using an LLM-based evaluation method.
A recently introduced metric, A3CU~\cite{liu2023towards}, is used to measure content similarity, which has better performance than traditional metrics like ROUGE.

\subsubsection{Results}

The main evaluation results are reported in Figure~\ref{fig:gptcompare-llama}, where the models are compared against GPT-3.5 under GPT-4's evaluation. 
Reference-based evaluation results are in Table~\ref{tab:gptrank-llama}.
We found that:

\noindent (1) Models trained in high resource settings can outperform those in low resource settings, highlighting the benefit of abundant training data.

\noindent (2) The model trained with contrastive learning, BRIO.Llama-2, outperforms GPT-3.5 under GPT-4's evaluation, indicating that a smaller summarization model has the capacity to reach LLM-level performance \textit{under LLM-based evaluation}.

\begin{table}[t!]
\small
\centering
\addtolength{\tabcolsep}{-3.0pt} 
\begin{tabular}{lcccc}
\toprule
\textbf{System} & \textbf{ROUGE-1}  & \textbf{ROUGE-2} & \textbf{A3CU} & \textbf{Length} \\
\midrule
 Llama-2      & 100.00  & 100.00  & 95.54 & 93.9   \\
\midrule
 BART.Llama-2 & 58.96   & \textbf{37.39}   & 50.71 & 92.1   \\
 BRIO.Llama-2 & \textbf{60.55}   & 37.38   & \textbf{53.25} & 92.5   \\
\bottomrule
\end{tabular}
\addtolength{\tabcolsep}{+3.0pt} 
\caption{Reference-based evaluation results under the high resource setting on CNNDM.
\textbf{BART.Llama-2} is fine-tuned with MLE training while \textbf{BRIO.Llama-2} is fine-tuned with contrastive learning.
}
\label{tab:gptrank-llama} 
\end{table}

\section{Human Evaluation and Meta-Analysis}
\label{sec:meta}

In \S\ref{sec:exp} we have demonstrated that smaller summarization models that are trained with better references can achieve on-par or even better performance than LLMs under \textbf{\textit{LLM-based evaluation}}.
However, the alignment between LLM and human evaluation still requires examination.
Therefore, we first conduct a human evaluation comparing the performance of models in \S\ref{sec:exp}, then perform a meta-analysis regarding the LLM-based evaluation.

\subsection{Human Evaluation Collection}
\label{subsec:human-eval-collection}

\paragraph{Evaluation Design} 
We formulate the human evaluation as a summary \textbf\textit{{pairwise}} comparison task.\footnote{The summary pairs are randomly shuffled.}
The summary pairs are compared on three aspects: (1) salience, (2) coherence, and (3) overall preference/quality, where the annotators are required to choose which summary is better (ties are allowed).
The detailed aspect definitions are in Appendix~\ref{subsec:aspect}.

\paragraph{Crowd-Annotation Collection}
We use Amazon Mechanical Turk\footnote{\url{https://www.mturk.com/}} (MTurk) for the crowd-annotation collection. 
Each data example is annotated by three annotators who are given two minutes for one task and compensated accordingly.
The participated crowd-annotators need to pass related qualification tests and have previous experience in evaluating summary quality.
We choose three system pairs for the collection on 100 test examples, where GPT-3.5 is the baseline LLM, and three BART checkpoints from \S\ref{subsec:exp-gptrank} are compared against GPT-3.5.
To check the inter-annotator agreement, we calculate the Krippendorff’s alpha~\cite{Krippendorff2011ComputingKA} with
MASI distance~\cite{passonneau-2006-measuring} following \citet{Goyal2022NewsSA}.
We found the average agreement to be 0.064, close to the agreement (0.05) reported by \citet{Goyal2022NewsSA} for similar evaluation settings.

\paragraph{Expert Evaluation}
The low agreement of crowd-annotation raises concerns about annotation quality.
Therefore, we (the co-authors)
conducted a careful expert evaluation to better understand this phenomenon and provide more trustworthy evaluation results.
We select 50 test examples to perform a pairwise comparison on three crowd-evaluated system groups and four additional groups.
We found the average agreement to be 0.044 among the expert annotators after a careful annotation, which re-confirms the hypotheses made in the related work~\cite{Goyal2022NewsSA, 10.1162/tacl_a_00632} regarding the inherent subjectivity of summarization evaluation especially when comparing summaries with similar quality. 
Besides, the experts agree with each other 55\% of the time, similar to the agreement level (65\%) in recent work~\cite{rafailov2023direct}.
We provide further analyses in Appendix \ref{subsec:annotation-examples}, which shows two main scenarios: (1) cases where the annotators unanimously favor LLM summaries; (2) cases where both LLM and smaller LM have good performance, resulting in different annotator preferences.
While higher agreement might be achieved with a more constrained evaluation protocol, we believe such a higher agreement can be ``artificial'' and cannot reflect the diverse distribution of human preferences.

\begin{table}[t!]
\small
\centering
\addtolength{\tabcolsep}{-2.0pt}
\begin{tabular}{clccc}
\toprule
\textbf{Group} & \textbf{System} & \textbf{Salience} & \textbf{Coherence} & \textbf{Overall} \\
\midrule
\multirow{2}{*}{1} & GPT-3.5 & 83 & 84 & 87 \\
&  BART & 26 & 34 & 20 \\
\midrule
\multirow{2}{*}{2} & GPT-3.5 & 68 & 68 & 62\\
&  BART.GPT-4 & 45 & 63 & 41 \\
\midrule
\multirow{2}{*}{3} & GPT-3.5 & 60 & 65 & 61 \\
&  BRIO.GPT-4 & 50 & 56 & 39 \\
\bottomrule
\end{tabular}
\addtolength{\tabcolsep}{2.0pt}
\caption{Crowd-annotations conducted on 3 groups of system pairs on 100 examples. The count of \textit{wins} for each system is reported, including ties as dual wins.
}
\label{tab:human-eval} 
\end{table}

\begin{table}[t!]
\small
\centering
\addtolength{\tabcolsep}{-2.5pt}
\begin{tabular}{clccc}
\toprule
\textbf{Group} & \textbf{System} & \textbf{Salience} & \textbf{Coherence} & \textbf{Overall} \\
\midrule
\multirow{2}{*}{1} & GPT-3.5 & 44 & 49 & 49 \\
&  BART & 10 & 4 & 2 \\
\midrule
\multirow{2}{*}{2} & GPT-3.5 & 40  & 35  & 35 \\
&  BART.GPT-4 & 22  &  24 &  18 \\
\midrule
\multirow{2}{*}{3} & GPT-3.5 & \textit{32}  & 39  & 33 \\
&  BRIO.GPT-4 & \textit{29}  &  24 &  21 \\
\midrule
\multirow{2}{*}{4} & BART.GPT-4 & 22 & 26 & 17\\
&  BRIO.GPT-4 & 41 & 36 & 39 \\
\midrule
\multirow{2}{*}{5} & GPT-3.5 & 36 & 38 & 38\\
&  BART.Llama-2 & 19 & 28 & 18  \\
\midrule
\multirow{2}{*}{6} & GPT-3.5 & 34 & 33 & \textit{28} \\
&  BRIO.Llama-2 &  22 & 33 & \textit{25} \\
\midrule
\multirow{2}{*}{7} & BART.Llama-2 & 25 & 28 & 22 \\
&  BRIO.Llama-2 &  31 & 36 & 31 \\
\bottomrule
\end{tabular}
\addtolength{\tabcolsep}{2.5pt}
\caption{Expert evaluation conducted on 7 groups of system pairs on 50 examples. 
The count of \textit{wins} for each system is reported, including ties as dual wins.
}
\label{tab:expert-eval} 
\end{table}

\subsection{Result Analysis}
\label{subsec:human-eval-results}

The crowd-annotation and expert-evaluation results are in Table~\ref{tab:human-eval} \& \ref{tab:expert-eval} respectively.
We note:

\noindent (1) The models trained with the LLMs as references can outperform the BART checkpoint trained on the original CNNDM dataset by a large margin, showing the importance of better references.

\noindent (2) When under a direct comparison in expert evaluation, BRIO.GPT-4/Llama-2 can outperform BART.GPT-4/Llama-2 on three aspects, demonstrating the effectiveness of contrastive learning.

\noindent (3)
While the smaller models cannot yet outperform GPT-3.5, the performance gap is smaller, with BRIO.GPT-4 achieving similar salience scores and BRIO.Llama-2 achieving similar overall scores.

\subsection{Meta-Analysis of LLM-based Evaluation}

BRIO.Llama-2 cannot outperform GPT-3.5 under human evaluation, even though they are favored by the evaluation methods based on GPT-4 (Figure~\ref{fig:gptcompare-llama}).
Therefore, we further investigate this discrepancy between human and LLM-based evaluation.

\paragraph{Human-LLM Alignment} We use the expert evaluation results to evaluate the performance of LLM-based evaluation as well as the crowd-annotation, by computing their agreements with the majority vote of expert evaluation on evaluation group 2 and 3 in Table~\ref{tab:human-eval} \& \ref{tab:expert-eval}.
Apart from GPTScore and GPTRank, we also compare the performance of G-Eval~\cite{liu-etal-2023-g}.
The prompts used for GPTRank and G-Eval are aspect-specific.
More details are in Appendix \ref{subsec:llm-metric-setting}.
The agreements are reported in Table~\ref{tab:meta-eval}, showing the following trends:

\noindent (1) LLM-based evaluation methods vary in performance, and GPT-4 outperforms GPT-3.5.

\noindent (2) GPT-4 with GPTRank can already outperform the performance of individual crowd-workers, while majority voting from crowd-workers still achieves the highest agreement.

\paragraph{LLM Positional Bias and Self-Inconsistency}
The GPTRank evaluation protocol performs pairwise comparisons, with which the LLMs can have a positional bias favoring either the first output or the second in the comparison~\citep{wang2023large}.
We observe that both GPT-3.5 and GPT-4 have similar positional biases in our study, which lead to a self-inconsistency -- the LLMs can favor different outputs when the output order is flipped in the pairwise comparison.
In Table~\ref{tab:incosistency}, we highlight this positional bias and the self-inconsistency rate of the LLMs.
Both LLMs have a bias toward the second output, and GPT-4 gave inconsistency decisions around 50\% of the time when the order of two outputs is flipped.

\paragraph{Discussion}
Our meta-analysis reveals that LLMs are \textit{noisy} summary evaluators because of the relatively low alignment level with human evaluation and the self-inconsistency.
We note that they are still better references for smaller model training compared with the original reference summaries (\S\ref{subsec:human-eval-results}).
However, we advise against using \textit{only} LLMs for system evaluation, particularly when comparing closely performing systems.

\begin{table}[t!]
\small
\centering
\addtolength{\tabcolsep}{-1pt}
\begin{tabular}{lrrr}
\toprule
   & \textbf{Salience} & \textbf{Coherence} & \textbf{Overall} \\
\midrule
Crowd-Individual & 0.189 & 0.061 & 0.062 \\
Crowd-Major-Voting & 0.241 & 0.116 & 0.166 \\
\midrule
G-EVAL-3.5 & -0.214 & -0.168 & -0.114 \\
G-EVAL-4 & -0.082 & -0.143 & -0.019 \\
GPTScore & -0.115 & -0.021 & -0.029   \\
GPT-3.5Rank & 0.036 & -0.034 & 0.018 \\
GPT-4Rank & 0.191 & 0.051 & 0.105  \\
\bottomrule
\end{tabular}
\addtolength{\tabcolsep}{1pt}
\caption{Performance comparison of LLM-based evaluation and crowd-annotation in terms of their agreements with expert evaluation. 
G-EVAL-3.5 and G-Eval-4 are G-Eval scores based on GPT-3.5 and GPT-4 respectively.
GPTScore is based on GPT3D3.
GPT-3.5Rank and GPT-4Rank are two versions of the GPTRank.
}
\label{tab:meta-eval} 
\end{table}

\begin{table}[t!]
\small
\centering
\begin{tabular}{lrrr}
\toprule
   & \textbf{Output 1} & \textbf{Output 2} & \textbf{Inconsistency} \\
\midrule
GPT-3.5 & 33.70\% & 66.30\% & 45.16\% \\
GPT-4 & 26.44\% & 73.56\% & 50.96\% \\
\bottomrule
\end{tabular}
\caption{Positional bias and self-inconsistency rate of GPTRank with GPT-3.5 and GPT-4 as the backbone models (ties are ignored).
Both LLMs have a bias toward the second output in pairwise comparisons.
}
\label{tab:incosistency} 
\end{table}

\section{Related Work}

\paragraph{Training Methods of Text Generation Models}
The standard MLE training of text generation models has two major limitations: 
(1) a discrepancy between the training objective, i.e., the cross-entropy loss, and the evaluation criteria (e.g., ROUGE);
(2) a discrepancy between the teacher-forcing~\citep{10.1162/neco.1989.1.2.270} training manner and auto-regressive generation behavior during evaluation, which is known as the \textit{exposure bias}~\citep{10.5555/2969239.2969370, DBLP:journals/corr/RanzatoCAZ15}.
As a result, training methods beyond MLE have been proposed to address these two limitations.
Among them a family of methods is based on \textit{reinforcement learning} (RL), which can optimize the text generation model toward a specific reward function~\citep{DBLP:journals/corr/RanzatoCAZ15, bahdanau2017an, li-etal-2016-deep, paulus2018a, li-etal-2019-deep, NEURIPS2020_1f89885d, pang2021text}.
Apart from RL, training methods based on supervised learning have also been developed, such as Minimum Risk Training~\citep{shen-etal-2016-minimum, wieting-etal-2019-beyond}, targeting a sequence-level optimization with various reward signals~\citep{wiseman-rush-2016-sequence, edunov-etal-2018-classical}.
More recently, \textit{contrastive learning}~\citep{10.1109/CVPR.2006.100} has also been adopted, which enhances the model ability by requiring the model to differentiate positive (good) and negative (bad) examples~\citep{yang-etal-2019-reducing, pan-etal-2021-contrastive, cao-wang-2021-cliff, liu-liu-2021-simcls, DBLP:journals/corr/abs-2108-11846, zhao2023calibrating, Zhang2022MomentumCF}. 
The latest work along this path has explored using contrastive learning to align LLMs with human feedback~\citep{yuan2023rrhf, Zhao2023SLiCHFSL, rafailov2023direct}, an alternative to reinforcement learning with human feedback~\cite{NEURIPS2020_1f89885d, NEURIPS2022_b1efde53}.

\paragraph{LLM-based Automatic Evaluation}
Recent work has explored using LLMs for automatic NLP evaluation.
GPTScore~\citep{Fu2023GPTScoreEA} leverages the LLM-predicted probability of text sequences as the quality score.
On the other hand, a line of work~\citep{chiang-lee-2023-large, Gao2023HumanlikeSE, chen-etal-2023-exploring-use, wang-etal-2023-chatgpt, Luo2023ChatGPTAA}, e.g., G-Eval~\citep{liu-etal-2023-g}, proposes evaluation methods that use LLMs to perform text completion tasks, such as predicting the answer of a Likert scale evaluation or pairwise comparison. 
Notably, several of these studies~\cite {Fu2023GPTScoreEA, liu-etal-2023-g, zheng2023judging, dubois2023alpacafarm, Gao2023HumanlikeSE, chen-etal-2023-exploring-use, wang-etal-2023-chatgpt} all evaluate the LLM-based evaluation methods on SummEval~\cite{fabbri-etal-2021-summeval}, a summarization human evaluation benchmark, and found that LLM-based evaluation has a higher correlation with human judgments than previous methods such as ROUGE or BERTScore~\citep{Zhang*2020BERTScore:}.
Apart from summarization evaluation, LLM-based evaluation has also been used in text classification tasks~\cite{Gilardi2023ChatGPTOC} and for reward design for RL agents~\cite{kwon2023reward}.

\paragraph{LLM Distillation and LLM-based Data Augmentation}
To improve the performance of smaller NLP models, related work has proposed methods of distilling LLMs and using LLMs for data augmentation~\citep{wang-etal-2021-want-reduce, ding-etal-2023-gpt, kang-etal-2023-distill}. 
Specifically, a line of work~\citep{shridhar-etal-2023-distilling, LI2022ExplanationsFL, hsieh-etal-2023-distilling} uses LLMs to generate both final answers and task-related descriptions for training smaller models on reasoning tasks, and Orca~\cite{mukherjee2023orca} extends this method for LLM distillation by training smaller models on the LLM-generated explanations.
Regarding text summarization, \citet{wang-etal-2021-want-reduce} introduces using GPT-3~\cite{NEURIPS2020_1457c0d6} to generate reference summaries while \citet{gekhman-etal-2023-trueteacher} proposes using LLMs to annotate the summary factual consistency~\cite{maynez-etal-2020-faithfulness} for the training of smaller evaluation models.

\section{Conclusion}

In this work, we study a learning setting of text summarization models where the LLMs are set to be the reference.
For this setting, we leverage the LLM-based evaluation methods to guide the model training through contrastive learning and empirically demonstrate the efficiency and effectiveness of our methods. 
Furthermore, we conduct human evaluation and meta-analysis regarding the reliability of LLM-based evaluation, which reveals its benefits as better training references and its limitations in terms of the alignment with human evaluation.
We believe our findings shed light on the direction of reliably applying the LLMs to the entire development loop (i.e., training-validation-evaluation) of smaller, task-specific NLP models, which has the potential to provide a balance between model performance and computational cost.

\section{Limitations}
The LLM-based evaluation results we reported are from OpenAI's APIs, which are subject to change.
Therefore, the reproducibility of our experiments is limited. 
To mitigate this problem, we will release the training data, model outputs, and LLM and human evaluation results to facilitate future work.

Both the LLM-based and human evaluations we conducted can be resource-intensive, requiring substantial time and budget.
As a result, we try to find a balance between the reliability of the evaluation result and the constraints of time and budget when selecting the sample size we used for evaluation. 
An evaluation at a larger scale is likely to yield more reliable results, which we leave for more dedicated future work in this direction.
The resource constraints also led us to use news summarization as a case study, leaving other summarization task scenarios for future work.

We chose not to include summary factual consistency as an individual quality aspect in human evaluation and the meta-analysis of LLM-based evaluation.
Related work~\cite{tang-etal-2023-understanding, 10.1162/tacl_a_00632} has found that the factual error rate is low on CNNDM dataset, especially for LLM summaries.
During our expert evaluation, the authors also did not observe significant flaws in factual consistency.
As a result, it would require a much larger sample size for an evaluation of factual consistency in order to understand the error patterns, which is out of the scope of this work.
However, we believe that such an evaluation is important for better understanding the summary quality of LLMs and LLM-supervised models, and we hope that the outcome of this work (e.g., the system outputs) can be a helpful resource for future work on this topic.

\section*{Acknowledgements}
We thank the anonymous reviewers for their constructive comments.
We are grateful for the compute support provided by the Google TRC program.

\bibliography{anthology,custom}

\begin{thebibliography}{72}
\expandafter\ifx\csname natexlab\endcsname\relax\def\natexlab#1{#1}\fi

\bibitem[{Bahdanau et~al.(2017)Bahdanau, Brakel, Xu, Goyal, Lowe, Pineau, Courville, and Bengio}]{bahdanau2017an}
Dzmitry Bahdanau, Philemon Brakel, Kelvin Xu, Anirudh Goyal, Ryan Lowe, Joelle Pineau, Aaron Courville, and Yoshua Bengio. 2017.
\newblock \href {https://openreview.net/forum?id=SJDaqqveg} {An actor-critic algorithm for sequence prediction}.
\newblock In \emph{International Conference on Learning Representations}.

\bibitem[{Bengio et~al.(2015)Bengio, Vinyals, Jaitly, and Shazeer}]{10.5555/2969239.2969370}
Samy Bengio, Oriol Vinyals, Navdeep Jaitly, and Noam Shazeer. 2015.
\newblock Scheduled sampling for sequence prediction with recurrent neural networks.
\newblock In \emph{Proceedings of the 28th International Conference on Neural Information Processing Systems - Volume 1}, NIPS'15, page 1171–1179, Cambridge, MA, USA. MIT Press.

\bibitem[{Brown et~al.(2020)Brown, Mann, Ryder, Subbiah, Kaplan, Dhariwal, Neelakantan, Shyam, Sastry, Askell, Agarwal, Herbert{-}Voss, Krueger, Henighan, Child, Ramesh, Ziegler, Wu, Winter, Hesse, Chen, Sigler, Litwin, Gray, Chess, Clark, Berner, McCandlish, Radford, Sutskever, and Amodei}]{NEURIPS2020_1457c0d6}
Tom~B. Brown, Benjamin Mann, Nick Ryder, Melanie Subbiah, Jared Kaplan, Prafulla Dhariwal, Arvind Neelakantan, Pranav Shyam, Girish Sastry, Amanda Askell, Sandhini Agarwal, Ariel Herbert{-}Voss, Gretchen Krueger, Tom Henighan, Rewon Child, Aditya Ramesh, Daniel~M. Ziegler, Jeffrey Wu, Clemens Winter, Christopher Hesse, Mark Chen, Eric Sigler, Mateusz Litwin, Scott Gray, Benjamin Chess, Jack Clark, Christopher Berner, Sam McCandlish, Alec Radford, Ilya Sutskever, and Dario Amodei. 2020.
\newblock \href {https://proceedings.neurips.cc/paper/2020/hash/1457c0d6bfcb4967418bfb8ac142f64a-Abstract.html} {Language models are few-shot learners}.
\newblock In \emph{Advances in Neural Information Processing Systems 33: Annual Conference on Neural Information Processing Systems 2020, NeurIPS 2020, December 6-12, 2020, virtual}.

\bibitem[{Cao and Wang(2021)}]{cao-wang-2021-cliff}
Shuyang Cao and Lu~Wang. 2021.
\newblock \href {https://doi.org/10.18653/v1/2021.emnlp-main.532} {{CLIFF}: Contrastive learning for improving faithfulness and factuality in abstractive summarization}.
\newblock In \emph{Proceedings of the 2021 Conference on Empirical Methods in Natural Language Processing}, pages 6633--6649, Online and Punta Cana, Dominican Republic. Association for Computational Linguistics.

\bibitem[{Chen et~al.(2023)Chen, Wang, Jiang, Shi, and Xu}]{chen-etal-2023-exploring-use}
Yi~Chen, Rui Wang, Haiyun Jiang, Shuming Shi, and Ruifeng Xu. 2023.
\newblock \href {https://doi.org/10.18653/v1/2023.findings-ijcnlp.32} {Exploring the use of large language models for reference-free text quality evaluation: An empirical study}.
\newblock In \emph{Findings of the Association for Computational Linguistics: IJCNLP-AACL 2023 (Findings)}, pages 361--374, Nusa Dua, Bali. Association for Computational Linguistics.

\bibitem[{Chiang and Lee(2023)}]{chiang-lee-2023-large}
Cheng-Han Chiang and Hung-yi Lee. 2023.
\newblock \href {https://doi.org/10.18653/v1/2023.acl-long.870} {Can large language models be an alternative to human evaluations?}
\newblock In \emph{Proceedings of the 61st Annual Meeting of the Association for Computational Linguistics (Volume 1: Long Papers)}, pages 15607--15631, Toronto, Canada. Association for Computational Linguistics.

\bibitem[{Chung et~al.(2022)Chung, Hou, Longpre, Zoph, Tay, Fedus, Li, Wang, Dehghani, Brahma, Webson, Gu, Dai, Suzgun, Chen, Chowdhery, Valter, Narang, Mishra, Yu, Zhao, Huang, Dai, Yu, Petrov, hsin Chi, Dean, Devlin, Roberts, Zhou, Le, and Wei}]{Chung2022ScalingIL}
Hyung~Won Chung, Le~Hou, S.~Longpre, Barret Zoph, Yi~Tay, William Fedus, Eric Li, Xuezhi Wang, Mostafa Dehghani, Siddhartha Brahma, Albert Webson, Shixiang~Shane Gu, Zhuyun Dai, Mirac Suzgun, Xinyun Chen, Aakanksha Chowdhery, Dasha Valter, Sharan Narang, Gaurav Mishra, Adams~Wei Yu, Vincent Zhao, Yanping Huang, Andrew~M. Dai, Hongkun Yu, Slav Petrov, Ed~Huai hsin Chi, Jeff Dean, Jacob Devlin, Adam Roberts, Denny Zhou, Quoc~V. Le, and Jason Wei. 2022.
\newblock Scaling instruction-finetuned language models.
\newblock \emph{ArXiv}, abs/2210.11416.

\bibitem[{Ding et~al.(2023)Ding, Qin, Liu, Chia, Li, Joty, and Bing}]{ding-etal-2023-gpt}
Bosheng Ding, Chengwei Qin, Linlin Liu, Yew~Ken Chia, Boyang Li, Shafiq Joty, and Lidong Bing. 2023.
\newblock \href {https://doi.org/10.18653/v1/2023.acl-long.626} {Is {GPT}-3 a good data annotator?}
\newblock In \emph{Proceedings of the 61st Annual Meeting of the Association for Computational Linguistics (Volume 1: Long Papers)}, pages 11173--11195, Toronto, Canada. Association for Computational Linguistics.

\bibitem[{Dubois et~al.(2023)Dubois, Li, Taori, Zhang, Gulrajani, Ba, Guestrin, Liang, and Hashimoto}]{dubois2023alpacafarm}
Yann Dubois, Xuechen Li, Rohan Taori, Tianyi Zhang, Ishaan Gulrajani, Jimmy Ba, Carlos Guestrin, Percy Liang, and Tatsunori Hashimoto. 2023.
\newblock \href {https://openreview.net/forum?id=4hturzLcKX} {Alpacafarm: A simulation framework for methods that learn from human feedback}.
\newblock In \emph{Thirty-seventh Conference on Neural Information Processing Systems}.

\bibitem[{Edunov et~al.(2018)Edunov, Ott, Auli, Grangier, and Ranzato}]{edunov-etal-2018-classical}
Sergey Edunov, Myle Ott, Michael Auli, David Grangier, and Marc{'}Aurelio Ranzato. 2018.
\newblock \href {https://doi.org/10.18653/v1/N18-1033} {Classical structured prediction losses for sequence to sequence learning}.
\newblock In \emph{Proceedings of the 2018 Conference of the North {A}merican Chapter of the Association for Computational Linguistics: Human Language Technologies, Volume 1 (Long Papers)}, pages 355--364, New Orleans, Louisiana. Association for Computational Linguistics.

\bibitem[{Fabbri et~al.(2021)Fabbri, Kry{\'s}ci{\'n}ski, McCann, Xiong, Socher, and Radev}]{fabbri-etal-2021-summeval}
Alexander~R. Fabbri, Wojciech Kry{\'s}ci{\'n}ski, Bryan McCann, Caiming Xiong, Richard Socher, and Dragomir Radev. 2021.
\newblock \href {https://doi.org/10.1162/tacl_a_00373} {{S}umm{E}val: Re-evaluating summarization evaluation}.
\newblock \emph{Transactions of the Association for Computational Linguistics}, 9:391--409.

\bibitem[{Fu et~al.(2023)Fu, Ng, Jiang, and Liu}]{Fu2023GPTScoreEA}
Jinlan Fu, See-Kiong Ng, Zhengbao Jiang, and Pengfei Liu. 2023.
\newblock {GPTScore}: Evaluate as you desire.
\newblock \emph{ArXiv}, abs/2302.04166.

\bibitem[{Gao et~al.(2023)Gao, Ruan, Sun, Yin, Yang, and Wan}]{Gao2023HumanlikeSE}
Mingqi Gao, Jie Ruan, Renliang Sun, Xunjian Yin, Shiping Yang, and Xiaojun Wan. 2023.
\newblock Human-like summarization evaluation with chatgpt.
\newblock \emph{ArXiv}, abs/2304.02554.

\bibitem[{Gehrmann et~al.(2021)Gehrmann, Adewumi, Aggarwal, Ammanamanchi, Aremu, Bosselut, Chandu, Clinciu, Das, Dhole, Du, Durmus, Du{\v{s}}ek, Emezue, Gangal, Garbacea, Hashimoto, Hou, Jernite, Jhamtani, Ji, Jolly, Kale, Kumar, Ladhak, Madaan, Maddela, Mahajan, Mahamood, Majumder, Martins, McMillan-Major, Mille, van Miltenburg, Nadeem, Narayan, Nikolaev, Niyongabo~Rubungo, Osei, Parikh, Perez-Beltrachini, Rao, Raunak, Rodriguez, Santhanam, Sedoc, Sellam, Shaikh, Shimorina, Sobrevilla~Cabezudo, Strobelt, Subramani, Xu, Yang, Yerukola, and Zhou}]{gehrmann-etal-2021-gem}
Sebastian Gehrmann, Tosin Adewumi, Karmanya Aggarwal, Pawan~Sasanka Ammanamanchi, Anuoluwapo Aremu, Antoine Bosselut, Khyathi~Raghavi Chandu, Miruna-Adriana Clinciu, Dipanjan Das, Kaustubh Dhole, Wanyu Du, Esin Durmus, Ond{\v{r}}ej Du{\v{s}}ek, Chris~Chinenye Emezue, Varun Gangal, Cristina Garbacea, Tatsunori Hashimoto, Yufang Hou, Yacine Jernite, Harsh Jhamtani, Yangfeng Ji, Shailza Jolly, Mihir Kale, Dhruv Kumar, Faisal Ladhak, Aman Madaan, Mounica Maddela, Khyati Mahajan, Saad Mahamood, Bodhisattwa~Prasad Majumder, Pedro~Henrique Martins, Angelina McMillan-Major, Simon Mille, Emiel van Miltenburg, Moin Nadeem, Shashi Narayan, Vitaly Nikolaev, Andre Niyongabo~Rubungo, Salomey Osei, Ankur Parikh, Laura Perez-Beltrachini, Niranjan~Ramesh Rao, Vikas Raunak, Juan~Diego Rodriguez, Sashank Santhanam, Jo{\~a}o Sedoc, Thibault Sellam, Samira Shaikh, Anastasia Shimorina, Marco~Antonio Sobrevilla~Cabezudo, Hendrik Strobelt, Nishant Subramani, Wei Xu, Diyi Yang, Akhila Yerukola, and Jiawei Zhou. 2021.
\newblock \href {https://doi.org/10.18653/v1/2021.gem-1.10} {The {GEM} benchmark: Natural language generation, its evaluation and metrics}.
\newblock In \emph{Proceedings of the 1st Workshop on Natural Language Generation, Evaluation, and Metrics (GEM 2021)}, pages 96--120, Online. Association for Computational Linguistics.

\bibitem[{Gehrmann et~al.(2022)Gehrmann, Bhattacharjee, Mahendiran, Wang, Papangelis, Madaan, McMillan-Major, Shvets, Upadhyay, Yao, Wilie, Bhagavatula, You, Thomson, Garbacea, Wang, Deutsch, Xiong, Jin, Gkatzia, Radev, Clark, Durmus, Ladhak, Ginter, Winata, Strobelt, Hayashi, Novikova, Kanerva, Chim, Zhou, Clive, Maynez, Sedoc, Juraska, Dhole, Chandu, Ribeiro, Tunstall, Zhang, Pushkarna, Creutz, White, Kale, Eddine, Daheim, Subramani, Dusek, Liang, Ammanamanchi, Zhu, Puduppully, Kriz, Shahriyar, Cardenas, Mahamood, Osei, Cahyawijaya, vStajner, Montella, Shailza, Jolly, Mille, Hasan, Shen, Adewumi, Raunak, Raheja, Nikolaev, Tsai, Jernite, Xu, Sang, Liu, and Hou}]{Gehrmann2022GEMv2MN}
Sebastian Gehrmann, Abhik Bhattacharjee, Abinaya Mahendiran, Alex Wang, Alexandros Papangelis, Aman Madaan, Angelina McMillan-Major, Anna Shvets, Ashish Upadhyay, Bingsheng Yao, Bryan Wilie, Chandra Bhagavatula, Chaobin You, Craig Thomson, Cristina Garbacea, Dakuo Wang, Daniel Deutsch, Deyi Xiong, Di~Jin, Dimitra Gkatzia, Dragomir~R. Radev, Elizabeth Clark, Esin Durmus, Faisal Ladhak, Filip Ginter, Genta~Indra Winata, Hendrik Strobelt, Hiroaki Hayashi, Jekaterina Novikova, Jenna Kanerva, Jenny Chim, Jiawei Zhou, Jordan Clive, Joshua Maynez, Jo{\~a}o Sedoc, Juraj Juraska, Kaustubh~D. Dhole, Khyathi~Raghavi Chandu, Leonardo F.~R. Ribeiro, Lewis Tunstall, Li~Zhang, Mahima Pushkarna, Mathias Creutz, Michael White, Mihir Kale, Moussa~Kamal Eddine, Nico Daheim, Nishant Subramani, Ondrej Dusek, Paul~Pu Liang, Pawan~Sasanka Ammanamanchi, Qinqin Zhu, Ratish Puduppully, Reno Kriz, Rifat Shahriyar, Ronald Cardenas, Saad Mahamood, Salomey Osei, Samuel Cahyawijaya, Sanja vStajner, S{\'e}bastien Montella, Shailza, Shailza
  Jolly, Simon Mille, Tahmid Hasan, Tianhao Shen, Tosin~P. Adewumi, Vikas Raunak, Vipul Raheja, Vitaly Nikolaev, Vivian Tsai, Yacine Jernite, Yi~Xu, Yisi Sang, Yixin Liu, and Yufang Hou. 2022.
\newblock Gemv2: Multilingual nlg benchmarking in a single line of code.
\newblock In \emph{Conference on Empirical Methods in Natural Language Processing}.

\bibitem[{Gekhman et~al.(2023)Gekhman, Herzig, Aharoni, Elkind, and Szpektor}]{gekhman-etal-2023-trueteacher}
Zorik Gekhman, Jonathan Herzig, Roee Aharoni, Chen Elkind, and Idan Szpektor. 2023.
\newblock \href {https://doi.org/10.18653/v1/2023.emnlp-main.127} {{T}rue{T}eacher: Learning factual consistency evaluation with large language models}.
\newblock In \emph{Proceedings of the 2023 Conference on Empirical Methods in Natural Language Processing}, pages 2053--2070, Singapore. Association for Computational Linguistics.

\bibitem[{Gilardi et~al.(2023)Gilardi, Alizadeh, and Kubli}]{Gilardi2023ChatGPTOC}
Fabrizio Gilardi, Meysam Alizadeh, and Ma{\"e}l Kubli. 2023.
\newblock Chatgpt outperforms crowd-workers for text-annotation tasks.
\newblock \emph{ArXiv}, abs/2303.15056.

\bibitem[{Goyal et~al.(2022)Goyal, Li, and Durrett}]{Goyal2022NewsSA}
Tanya Goyal, Junyi~Jessy Li, and Greg Durrett. 2022.
\newblock News summarization and evaluation in the era of gpt-3.
\newblock \emph{ArXiv}, abs/2209.12356.

\bibitem[{Hadsell et~al.(2006)Hadsell, Chopra, and LeCun}]{10.1109/CVPR.2006.100}
Raia Hadsell, Sumit Chopra, and Yann LeCun. 2006.
\newblock \href {https://doi.org/10.1109/CVPR.2006.100} {Dimensionality reduction by learning an invariant mapping}.
\newblock In \emph{Proceedings of the 2006 IEEE Computer Society Conference on Computer Vision and Pattern Recognition - Volume 2}, CVPR '06, page 1735–1742, USA. IEEE Computer Society.

\bibitem[{Holtzman et~al.(2020)Holtzman, Buys, Du, Forbes, and Choi}]{Holtzman2020The}
Ari Holtzman, Jan Buys, Li~Du, Maxwell Forbes, and Yejin Choi. 2020.
\newblock \href {https://openreview.net/forum?id=rygGQyrFvH} {The curious case of neural text degeneration}.
\newblock In \emph{International Conference on Learning Representations}.

\bibitem[{Hsieh et~al.(2023)Hsieh, Li, Yeh, Nakhost, Fujii, Ratner, Krishna, Lee, and Pfister}]{hsieh-etal-2023-distilling}
Cheng-Yu Hsieh, Chun-Liang Li, Chih-kuan Yeh, Hootan Nakhost, Yasuhisa Fujii, Alex Ratner, Ranjay Krishna, Chen-Yu Lee, and Tomas Pfister. 2023.
\newblock \href {https://doi.org/10.18653/v1/2023.findings-acl.507} {Distilling step-by-step! outperforming larger language models with less training data and smaller model sizes}.
\newblock In \emph{Findings of the Association for Computational Linguistics: ACL 2023}, pages 8003--8017, Toronto, Canada. Association for Computational Linguistics.

\bibitem[{Kang et~al.(2023)Kang, Xu, and Ritter}]{kang-etal-2023-distill}
Junmo Kang, Wei Xu, and Alan Ritter. 2023.
\newblock \href {https://doi.org/10.18653/v1/2023.acl-long.622} {Distill or annotate? cost-efficient fine-tuning of compact models}.
\newblock In \emph{Proceedings of the 61st Annual Meeting of the Association for Computational Linguistics (Volume 1: Long Papers)}, pages 11100--11119, Toronto, Canada. Association for Computational Linguistics.

\bibitem[{Kim and Rush(2016)}]{kim-rush-2016-sequence}
Yoon Kim and Alexander~M. Rush. 2016.
\newblock \href {https://doi.org/10.18653/v1/D16-1139} {Sequence-level knowledge distillation}.
\newblock In \emph{Proceedings of the 2016 Conference on Empirical Methods in Natural Language Processing}, pages 1317--1327, Austin, Texas. Association for Computational Linguistics.

\bibitem[{Krippendorff(2011)}]{Krippendorff2011ComputingKA}
Klaus Krippendorff. 2011.
\newblock Computing krippendorff’s alpha-reliability.

\bibitem[{Kwon et~al.(2023)Kwon, Xie, Bullard, and Sadigh}]{kwon2023reward}
Minae Kwon, Sang~Michael Xie, Kalesha Bullard, and Dorsa Sadigh. 2023.
\newblock \href {https://openreview.net/forum?id=10uNUgI5Kl} {Reward design with language models}.
\newblock In \emph{The Eleventh International Conference on Learning Representations}.

\bibitem[{Lewis et~al.(2020)Lewis, Liu, Goyal, Ghazvininejad, Mohamed, Levy, Stoyanov, and Zettlemoyer}]{lewis-etal-2020-bart}
Mike Lewis, Yinhan Liu, Naman Goyal, Marjan Ghazvininejad, Abdelrahman Mohamed, Omer Levy, Veselin Stoyanov, and Luke Zettlemoyer. 2020.
\newblock \href {https://doi.org/10.18653/v1/2020.acl-main.703} {{BART}: Denoising sequence-to-sequence pre-training for natural language generation, translation, and comprehension}.
\newblock In \emph{Proceedings of the 58th Annual Meeting of the Association for Computational Linguistics}, pages 7871--7880, Online. Association for Computational Linguistics.

\bibitem[{Li et~al.(2016)Li, Monroe, Ritter, Jurafsky, Galley, and Gao}]{li-etal-2016-deep}
Jiwei Li, Will Monroe, Alan Ritter, Dan Jurafsky, Michel Galley, and Jianfeng Gao. 2016.
\newblock \href {https://doi.org/10.18653/v1/D16-1127} {Deep reinforcement learning for dialogue generation}.
\newblock In \emph{Proceedings of the 2016 Conference on Empirical Methods in Natural Language Processing}, pages 1192--1202, Austin, Texas. Association for Computational Linguistics.

\bibitem[{Li et~al.(2022)Li, Chen, Shen, Chen, Zhang, Li, Wang, Qian, Peng, Mao, Chen, and Yan}]{LI2022ExplanationsFL}
Shiyang Li, Jianshu Chen, Yelong Shen, Zhiyu Chen, Xinlu Zhang, Zekun Li, Hong Wang, Jingu Qian, Baolin Peng, Yi~Mao, Wenhu Chen, and Xifeng Yan. 2022.
\newblock Explanations from large language models make small reasoners better.
\newblock \emph{ArXiv}, abs/2210.06726.

\bibitem[{Li et~al.(2019)Li, Lei, Qin, and Wang}]{li-etal-2019-deep}
Siyao Li, Deren Lei, Pengda Qin, and William~Yang Wang. 2019.
\newblock \href {https://doi.org/10.18653/v1/D19-1623} {Deep reinforcement learning with distributional semantic rewards for abstractive summarization}.
\newblock In \emph{Proceedings of the 2019 Conference on Empirical Methods in Natural Language Processing and the 9th International Joint Conference on Natural Language Processing (EMNLP-IJCNLP)}, pages 6038--6044, Hong Kong, China. Association for Computational Linguistics.

\bibitem[{Liang et~al.(2023)Liang, Bommasani, Lee, Tsipras, Soylu, Yasunaga, Zhang, Narayanan, Wu, Kumar, Newman, Yuan, Yan, Zhang, Cosgrove, Manning, Re, Acosta-Navas, Hudson, Zelikman, Durmus, Ladhak, Rong, Ren, Yao, WANG, Santhanam, Orr, Zheng, Yuksekgonul, Suzgun, Kim, Guha, Chatterji, Khattab, Henderson, Huang, Chi, Xie, Santurkar, Ganguli, Hashimoto, Icard, Zhang, Chaudhary, Wang, Li, Mai, Zhang, and Koreeda}]{liang2023holistic}
Percy Liang, Rishi Bommasani, Tony Lee, Dimitris Tsipras, Dilara Soylu, Michihiro Yasunaga, Yian Zhang, Deepak Narayanan, Yuhuai Wu, Ananya Kumar, Benjamin Newman, Binhang Yuan, Bobby Yan, Ce~Zhang, Christian~Alexander Cosgrove, Christopher~D Manning, Christopher Re, Diana Acosta-Navas, Drew~Arad Hudson, Eric Zelikman, Esin Durmus, Faisal Ladhak, Frieda Rong, Hongyu Ren, Huaxiu Yao, Jue WANG, Keshav Santhanam, Laurel Orr, Lucia Zheng, Mert Yuksekgonul, Mirac Suzgun, Nathan Kim, Neel Guha, Niladri~S. Chatterji, Omar Khattab, Peter Henderson, Qian Huang, Ryan~Andrew Chi, Sang~Michael Xie, Shibani Santurkar, Surya Ganguli, Tatsunori Hashimoto, Thomas Icard, Tianyi Zhang, Vishrav Chaudhary, William Wang, Xuechen Li, Yifan Mai, Yuhui Zhang, and Yuta Koreeda. 2023.
\newblock \href {https://openreview.net/forum?id=iO4LZibEqW} {Holistic evaluation of language models}.
\newblock \emph{Transactions on Machine Learning Research}.
\newblock Featured Certification, Expert Certification.

\bibitem[{Lin(2004)}]{lin-2004-rouge}
Chin-Yew Lin. 2004.
\newblock \href {https://aclanthology.org/W04-1013} {{ROUGE}: A package for automatic evaluation of summaries}.
\newblock In \emph{Text Summarization Branches Out}, pages 74--81, Barcelona, Spain. Association for Computational Linguistics.

\bibitem[{Liu et~al.(2023{\natexlab{a}})Liu, Iter, Xu, Wang, Xu, and Zhu}]{liu-etal-2023-g}
Yang Liu, Dan Iter, Yichong Xu, Shuohang Wang, Ruochen Xu, and Chenguang Zhu. 2023{\natexlab{a}}.
\newblock \href {https://doi.org/10.18653/v1/2023.emnlp-main.153} {{G}-eval: {NLG} evaluation using gpt-4 with better human alignment}.
\newblock In \emph{Proceedings of the 2023 Conference on Empirical Methods in Natural Language Processing}, pages 2511--2522, Singapore. Association for Computational Linguistics.

\bibitem[{Liu et~al.(2024)Liu, Fabbri, Chen, Zhao, Han, Joty, Liu, Radev, Wu, and Cohan}]{liu2023benchmarking}
Yixin Liu, Alexander~R Fabbri, Jiawen Chen, Yilun Zhao, Simeng Han, Shafiq Joty, Pengfei Liu, Dragomir Radev, Chien-Sheng Wu, and Arman Cohan. 2024.
\newblock Benchmarking generation and evaluation capabilities of large language models for instruction controllable summarization.
\newblock In \emph{Findings of the Association for Computational Linguistics: NAACL 2024}.

\bibitem[{Liu et~al.(2023{\natexlab{b}})Liu, Fabbri, Liu, Zhao, Nan, Han, Han, Joty, Wu, Xiong, and Radev}]{Liu2022RevisitingTG}
Yixin Liu, Alexander~R. Fabbri, Pengfei Liu, Yilun Zhao, Linyong Nan, Ruilin Han, Simeng Han, Shafiq~R. Joty, Chien-Sheng Wu, Caiming Xiong, and Dragomir~R. Radev. 2023{\natexlab{b}}.
\newblock Revisiting the gold standard: Grounding summarization evaluation with robust human evaluation.
\newblock In \emph{Proceedings of the 61th Annual Meeting of the Association for Computational Linguistics}.

\bibitem[{Liu et~al.(2023{\natexlab{c}})Liu, Fabbri, Zhao, Liu, Joty, Wu, Xiong, and Radev}]{liu2023towards}
Yixin Liu, Alexander~R Fabbri, Yilun Zhao, Pengfei Liu, Shafiq Joty, Chien-Sheng Wu, Caiming Xiong, and Dragomir Radev. 2023{\natexlab{c}}.
\newblock Towards interpretable and efficient automatic reference-based summarization evaluation.
\newblock \emph{Proceedings of the 2023 Conference on Empirical Methods in Natural Language Processing}.

\bibitem[{Liu and Liu(2021)}]{liu-liu-2021-simcls}
Yixin Liu and Pengfei Liu. 2021.
\newblock \href {https://doi.org/10.18653/v1/2021.acl-short.135} {{S}im{CLS}: A simple framework for contrastive learning of abstractive summarization}.
\newblock In \emph{Proceedings of the 59th Annual Meeting of the Association for Computational Linguistics and the 11th International Joint Conference on Natural Language Processing (Volume 2: Short Papers)}, pages 1065--1072, Online. Association for Computational Linguistics.

\bibitem[{Liu et~al.(2022)Liu, Liu, Radev, and Neubig}]{liu-etal-2022-brio}
Yixin Liu, Pengfei Liu, Dragomir Radev, and Graham Neubig. 2022.
\newblock \href {https://doi.org/10.18653/v1/2022.acl-long.207} {{BRIO}: Bringing order to abstractive summarization}.
\newblock In \emph{Proceedings of the 60th Annual Meeting of the Association for Computational Linguistics (Volume 1: Long Papers)}, pages 2890--2903, Dublin, Ireland. Association for Computational Linguistics.

\bibitem[{Luo et~al.(2023)Luo, Xie, and Ananiadou}]{Luo2023ChatGPTAA}
Zheheng Luo, Qianqian Xie, and Sophia Ananiadou. 2023.
\newblock {ChatGPT} as a factual inconsistency evaluator for text summarization.
\newblock \emph{ArXiv}.

\bibitem[{Maynez et~al.(2020)Maynez, Narayan, Bohnet, and McDonald}]{maynez-etal-2020-faithfulness}
Joshua Maynez, Shashi Narayan, Bernd Bohnet, and Ryan McDonald. 2020.
\newblock \href {https://doi.org/10.18653/v1/2020.acl-main.173} {On faithfulness and factuality in abstractive summarization}.
\newblock In \emph{Proceedings of the 58th Annual Meeting of the Association for Computational Linguistics}, pages 1906--1919, Online. Association for Computational Linguistics.

\bibitem[{Mukherjee et~al.(2023)Mukherjee, Mitra, Jawahar, Agarwal, Palangi, and Awadallah}]{mukherjee2023orca}
Subhabrata Mukherjee, Arindam Mitra, Ganesh Jawahar, Sahaj Agarwal, Hamid Palangi, and Ahmed Awadallah. 2023.
\newblock Orca: Progressive learning from complex explanation traces of gpt-4.
\newblock \emph{arXiv preprint arXiv:2306.02707}.

\bibitem[{Nallapati et~al.(2016)Nallapati, Zhou, dos Santos, Gulcehre, and Xiang}]{Nallapati:16}
Ramesh Nallapati, Bowen Zhou, Cicero dos Santos, Caglar Gulcehre, and Bing Xiang. 2016.
\newblock \href {https://doi.org/10.18653/v1/K16-1028} {Abstractive text summarization using sequence-to-sequence {RNN}s and beyond}.
\newblock In \emph{Proceedings of The 20th {SIGNLL} Conference on Computational Natural Language Learning}, pages 280--290, Berlin, Germany. Association for Computational Linguistics.

\bibitem[{Narayan et~al.(2018)Narayan, Cohen, and Lapata}]{narayan-etal-2018-dont}
Shashi Narayan, Shay~B. Cohen, and Mirella Lapata. 2018.
\newblock \href {https://doi.org/10.18653/v1/D18-1206} {Don{'}t give me the details, just the summary! topic-aware convolutional neural networks for extreme summarization}.
\newblock In \emph{Proceedings of the 2018 Conference on Empirical Methods in Natural Language Processing}, pages 1797--1807, Brussels, Belgium. Association for Computational Linguistics.

\bibitem[{OpenAI(2023)}]{OpenAI2023GPT4TR}
OpenAI. 2023.
\newblock {GPT-4} technical report.
\newblock \emph{ArXiv}, abs/2303.08774.

\bibitem[{Ouyang et~al.(2022)Ouyang, Wu, Jiang, Almeida, Wainwright, Mishkin, Zhang, Agarwal, Slama, Ray, Schulman, Hilton, Kelton, Miller, Simens, Askell, Welinder, Christiano, Leike, and Lowe}]{NEURIPS2022_b1efde53}
Long Ouyang, Jeffrey Wu, Xu~Jiang, Diogo Almeida, Carroll Wainwright, Pamela Mishkin, Chong Zhang, Sandhini Agarwal, Katarina Slama, Alex Ray, John Schulman, Jacob Hilton, Fraser Kelton, Luke Miller, Maddie Simens, Amanda Askell, Peter Welinder, Paul~F Christiano, Jan Leike, and Ryan Lowe. 2022.
\newblock \href {https://proceedings.neurips.cc/paper_files/paper/2022/file/b1efde53be364a73914f58805a001731-Paper-Conference.pdf} {Training language models to follow instructions with human feedback}.
\newblock In \emph{Advances in Neural Information Processing Systems}, volume~35, pages 27730--27744. Curran Associates, Inc.

\bibitem[{Pan et~al.(2021)Pan, Wang, Wu, and Li}]{pan-etal-2021-contrastive}
Xiao Pan, Mingxuan Wang, Liwei Wu, and Lei Li. 2021.
\newblock \href {https://doi.org/10.18653/v1/2021.acl-long.21} {Contrastive learning for many-to-many multilingual neural machine translation}.
\newblock In \emph{Proceedings of the 59th Annual Meeting of the Association for Computational Linguistics and the 11th International Joint Conference on Natural Language Processing (Volume 1: Long Papers)}, pages 244--258, Online. Association for Computational Linguistics.

\bibitem[{Pang and He(2021)}]{pang2021text}
Richard~Yuanzhe Pang and He~He. 2021.
\newblock \href {https://openreview.net/forum?id=RovX-uQ1Hua} {Text generation by learning from demonstrations}.
\newblock In \emph{International Conference on Learning Representations}.

\bibitem[{Passonneau(2006)}]{passonneau-2006-measuring}
Rebecca Passonneau. 2006.
\newblock \href {http://www.lrec-conf.org/proceedings/lrec2006/pdf/636_pdf.pdf} {Measuring agreement on set-valued items ({MASI}) for semantic and pragmatic annotation}.
\newblock In \emph{Proceedings of the Fifth International Conference on Language Resources and Evaluation ({LREC}{'}06)}, Genoa, Italy. European Language Resources Association (ELRA).

\bibitem[{Paulus et~al.(2018)Paulus, Xiong, and Socher}]{paulus2018a}
Romain Paulus, Caiming Xiong, and Richard Socher. 2018.
\newblock \href {https://openreview.net/forum?id=HkAClQgA-} {A deep reinforced model for abstractive summarization}.
\newblock In \emph{International Conference on Learning Representations}.

\bibitem[{Pu et~al.(2023)Pu, Gao, and Wan}]{Pu2023SummarizationI}
Xiao Pu, Mingqi Gao, and Xiaojun Wan. 2023.
\newblock \href {https://api.semanticscholar.org/CorpusID:262044218} {Summarization is (almost) dead}.
\newblock \emph{ArXiv}, abs/2309.09558.

\bibitem[{Rafailov et~al.(2023)Rafailov, Sharma, Mitchell, Manning, Ermon, and Finn}]{rafailov2023direct}
Rafael Rafailov, Archit Sharma, Eric Mitchell, Christopher~D Manning, Stefano Ermon, and Chelsea Finn. 2023.
\newblock \href {https://openreview.net/forum?id=HPuSIXJaa9} {Direct preference optimization: Your language model is secretly a reward model}.
\newblock In \emph{Thirty-seventh Conference on Neural Information Processing Systems}.

\bibitem[{Rajani et~al.(2023)Rajani, Lambert, Han, Wang, Nitski, Beeching, and Tunstall}]{rajani2023llm_labels}
Nazneen Rajani, Nathan Lambert, Sheon Han, Jean Wang, Osvald Nitski, Edward Beeching, and Lewis Tunstall. 2023.
\newblock Can foundation models label data like humans?
\newblock \emph{Hugging Face Blog}.
\newblock Https://huggingface.co/blog/llm-v-human-data.

\bibitem[{Ranzato et~al.(2016)Ranzato, Chopra, Auli, and Zaremba}]{DBLP:journals/corr/RanzatoCAZ15}
Marc'Aurelio Ranzato, Sumit Chopra, Michael Auli, and Wojciech Zaremba. 2016.
\newblock \href {http://arxiv.org/abs/1511.06732} {Sequence level training with recurrent neural networks}.
\newblock In \emph{4th International Conference on Learning Representations, {ICLR} 2016, San Juan, Puerto Rico, May 2-4, 2016, Conference Track Proceedings}.

\bibitem[{Shen et~al.(2016)Shen, Cheng, He, He, Wu, Sun, and Liu}]{shen-etal-2016-minimum}
Shiqi Shen, Yong Cheng, Zhongjun He, Wei He, Hua Wu, Maosong Sun, and Yang Liu. 2016.
\newblock \href {https://doi.org/10.18653/v1/P16-1159} {Minimum risk training for neural machine translation}.
\newblock In \emph{Proceedings of the 54th Annual Meeting of the Association for Computational Linguistics (Volume 1: Long Papers)}, pages 1683--1692, Berlin, Germany. Association for Computational Linguistics.

\bibitem[{Shridhar et~al.(2023)Shridhar, Stolfo, and Sachan}]{shridhar-etal-2023-distilling}
Kumar Shridhar, Alessandro Stolfo, and Mrinmaya Sachan. 2023.
\newblock \href {https://doi.org/10.18653/v1/2023.findings-acl.441} {Distilling reasoning capabilities into smaller language models}.
\newblock In \emph{Findings of the Association for Computational Linguistics: ACL 2023}, pages 7059--7073, Toronto, Canada. Association for Computational Linguistics.

\bibitem[{Stiennon et~al.(2020)Stiennon, Ouyang, Wu, Ziegler, Lowe, Voss, Radford, Amodei, and Christiano}]{NEURIPS2020_1f89885d}
Nisan Stiennon, Long Ouyang, Jeffrey Wu, Daniel Ziegler, Ryan Lowe, Chelsea Voss, Alec Radford, Dario Amodei, and Paul~F Christiano. 2020.
\newblock \href {https://proceedings.neurips.cc/paper_files/paper/2020/file/1f89885d556929e98d3ef9b86448f951-Paper.pdf} {Learning to summarize with human feedback}.
\newblock In \emph{Advances in Neural Information Processing Systems}, volume~33, pages 3008--3021. Curran Associates, Inc.

\bibitem[{Sun and Li(2021)}]{DBLP:journals/corr/abs-2108-11846}
Shichao Sun and Wenjie Li. 2021.
\newblock \href {http://arxiv.org/abs/2108.11846} {Alleviating exposure bias via contrastive learning for abstractive text summarization}.
\newblock \emph{CoRR}, abs/2108.11846.

\bibitem[{Tang et~al.(2023)Tang, Goyal, Fabbri, Laban, Xu, Yavuz, Kryscinski, Rousseau, and Durrett}]{tang-etal-2023-understanding}
Liyan Tang, Tanya Goyal, Alex Fabbri, Philippe Laban, Jiacheng Xu, Semih Yavuz, Wojciech Kryscinski, Justin Rousseau, and Greg Durrett. 2023.
\newblock \href {https://doi.org/10.18653/v1/2023.acl-long.650} {Understanding factual errors in summarization: Errors, summarizers, datasets, error detectors}.
\newblock In \emph{Proceedings of the 61st Annual Meeting of the Association for Computational Linguistics (Volume 1: Long Papers)}, pages 11626--11644, Toronto, Canada. Association for Computational Linguistics.

\bibitem[{Touvron et~al.(2023)Touvron, Martin, Stone, Albert, Almahairi, Babaei, Bashlykov, Batra, Bhargava, Bhosale et~al.}]{touvron2023llama}
Hugo Touvron, Louis Martin, Kevin Stone, Peter Albert, Amjad Almahairi, Yasmine Babaei, Nikolay Bashlykov, Soumya Batra, Prajjwal Bhargava, Shruti Bhosale, et~al. 2023.
\newblock Llama 2: Open foundation and fine-tuned chat models.
\newblock \emph{arXiv preprint arXiv:2307.09288}.

\bibitem[{Wang et~al.(2023{\natexlab{a}})Wang, Liang, Meng, Sun, Shi, Li, Xu, Qu, and Zhou}]{wang-etal-2023-chatgpt}
Jiaan Wang, Yunlong Liang, Fandong Meng, Zengkui Sun, Haoxiang Shi, Zhixu Li, Jinan Xu, Jianfeng Qu, and Jie Zhou. 2023{\natexlab{a}}.
\newblock \href {https://doi.org/10.18653/v1/2023.newsum-1.1} {Is {C}hat{GPT} a good {NLG} evaluator? a preliminary study}.
\newblock In \emph{Proceedings of the 4th New Frontiers in Summarization Workshop}, pages 1--11, Singapore. Association for Computational Linguistics.

\bibitem[{Wang et~al.(2023{\natexlab{b}})Wang, Li, Chen, Zhu, Lin, Cao, Liu, Liu, and Sui}]{wang2023large}
Peiyi Wang, Lei Li, Liang Chen, Dawei Zhu, Binghuai Lin, Yunbo Cao, Qi~Liu, Tianyu Liu, and Zhifang Sui. 2023{\natexlab{b}}.
\newblock Large language models are not fair evaluators.
\newblock \emph{arXiv preprint arXiv:2305.17926}.

\bibitem[{Wang et~al.(2021)Wang, Liu, Xu, Zhu, and Zeng}]{wang-etal-2021-want-reduce}
Shuohang Wang, Yang Liu, Yichong Xu, Chenguang Zhu, and Michael Zeng. 2021.
\newblock \href {https://doi.org/10.18653/v1/2021.findings-emnlp.354} {Want to reduce labeling cost? {GPT}-3 can help}.
\newblock In \emph{Findings of the Association for Computational Linguistics: EMNLP 2021}, pages 4195--4205, Punta Cana, Dominican Republic. Association for Computational Linguistics.

\bibitem[{Wieting et~al.(2019)Wieting, Berg-Kirkpatrick, Gimpel, and Neubig}]{wieting-etal-2019-beyond}
John Wieting, Taylor Berg-Kirkpatrick, Kevin Gimpel, and Graham Neubig. 2019.
\newblock \href {https://doi.org/10.18653/v1/P19-1427} {Beyond {BLEU}:training neural machine translation with semantic similarity}.
\newblock In \emph{Proceedings of the 57th Annual Meeting of the Association for Computational Linguistics}, pages 4344--4355, Florence, Italy. Association for Computational Linguistics.

\bibitem[{Williams and Zipser(1989)}]{10.1162/neco.1989.1.2.270}
Ronald~J. Williams and David Zipser. 1989.
\newblock \href {https://doi.org/10.1162/neco.1989.1.2.270} {A learning algorithm for continually running fully recurrent neural networks}.
\newblock \emph{Neural Comput.}, 1(2):270–280.

\bibitem[{Wiseman and Rush(2016)}]{wiseman-rush-2016-sequence}
Sam Wiseman and Alexander~M. Rush. 2016.
\newblock \href {https://doi.org/10.18653/v1/D16-1137} {Sequence-to-sequence learning as beam-search optimization}.
\newblock In \emph{Proceedings of the 2016 Conference on Empirical Methods in Natural Language Processing}, pages 1296--1306, Austin, Texas. Association for Computational Linguistics.

\bibitem[{Yang et~al.(2019)Yang, Cheng, Liu, and Sun}]{yang-etal-2019-reducing}
Zonghan Yang, Yong Cheng, Yang Liu, and Maosong Sun. 2019.
\newblock \href {https://doi.org/10.18653/v1/P19-1623} {Reducing word omission errors in neural machine translation: A contrastive learning approach}.
\newblock In \emph{Proceedings of the 57th Annual Meeting of the Association for Computational Linguistics}, pages 6191--6196, Florence, Italy. Association for Computational Linguistics.

\bibitem[{Yuan et~al.(2023)Yuan, Yuan, Tan, Wang, Huang, and Huang}]{yuan2023rrhf}
Hongyi Yuan, Zheng Yuan, Chuanqi Tan, Wei Wang, Songfang Huang, and Fei Huang. 2023.
\newblock \href {https://openreview.net/forum?id=EdIGMCHk4l} {{RRHF}: Rank responses to align language models with human feedback}.
\newblock In \emph{Thirty-seventh Conference on Neural Information Processing Systems}.

\bibitem[{Zhang* et~al.(2020)Zhang*, Kishore*, Wu*, Weinberger, and Artzi}]{Zhang*2020BERTScore:}
Tianyi Zhang*, Varsha Kishore*, Felix Wu*, Kilian~Q. Weinberger, and Yoav Artzi. 2020.
\newblock \href {https://openreview.net/forum?id=SkeHuCVFDr} {Bertscore: Evaluating text generation with bert}.
\newblock In \emph{International Conference on Learning Representations}.

\bibitem[{Zhang et~al.(2024)Zhang, Ladhak, Durmus, Liang, McKeown, and Hashimoto}]{10.1162/tacl_a_00632}
Tianyi Zhang, Faisal Ladhak, Esin Durmus, Percy Liang, Kathleen McKeown, and Tatsunori~B. Hashimoto. 2024.
\newblock \href {https://doi.org/10.1162/tacl_a_00632} {{Benchmarking Large Language Models for News Summarization}}.
\newblock \emph{Transactions of the Association for Computational Linguistics}, 12:39--57.

\bibitem[{Zhang et~al.(2022)Zhang, Liu, Wang, He, Yu, Chen, Xiong, and Wei}]{Zhang2022MomentumCF}
Xingxing Zhang, Yiran Liu, Xun Wang, Pengcheng He, Yang Yu, Si-Qing Chen, Wayne Xiong, and Furu Wei. 2022.
\newblock Momentum calibration for text generation.
\newblock \emph{ArXiv}, abs/2212.04257.

\bibitem[{Zhao et~al.(2023{\natexlab{a}})Zhao, Joshi, Liu, Khalman, Saleh, and Liu}]{Zhao2023SLiCHFSL}
Yao Zhao, Rishabh Joshi, Tianqi Liu, Misha Khalman, Mohammad Saleh, and Peter~J. Liu. 2023{\natexlab{a}}.
\newblock {SLiC-HF}: Sequence likelihood calibration with human feedback.
\newblock \emph{ArXiv}.

\bibitem[{Zhao et~al.(2023{\natexlab{b}})Zhao, Khalman, Joshi, Narayan, Saleh, and Liu}]{zhao2023calibrating}
Yao Zhao, Mikhail Khalman, Rishabh Joshi, Shashi Narayan, Mohammad Saleh, and Peter~J Liu. 2023{\natexlab{b}}.
\newblock \href {https://openreview.net/forum?id=0qSOodKmJaN} {Calibrating sequence likelihood improves conditional language generation}.
\newblock In \emph{The Eleventh International Conference on Learning Representations}.

\bibitem[{Zheng et~al.(2023)Zheng, Chiang, Sheng, Zhuang, Wu, Zhuang, Lin, Li, Li, Xing, Zhang, Gonzalez, and Stoica}]{zheng2023judging}
Lianmin Zheng, Wei-Lin Chiang, Ying Sheng, Siyuan Zhuang, Zhanghao Wu, Yonghao Zhuang, Zi~Lin, Zhuohan Li, Dacheng Li, Eric Xing, Hao Zhang, Joseph~E. Gonzalez, and Ion Stoica. 2023.
\newblock \href {https://openreview.net/forum?id=uccHPGDlao} {Judging {LLM}-as-a-judge with {MT}-bench and chatbot arena}.
\newblock In \emph{Thirty-seventh Conference on Neural Information Processing Systems Datasets and Benchmarks Track}.

\end{thebibliography}

\appendix

\section{Experimental Details}
\label{sec:appendix-exp}

\subsection{LLM Summary Generation}
\label{appendix:llm-gen}

In \S\ref{sec:exp}, we use the following prompt to generate the LLM summaries:

\begin{quote}
    Article: \{\{{Article}\}\}

Summarize the above article in three sentences.

Summary:
\end{quote}

Since text summarization is a conditional generation task that requires high accuracy, we set the sampling temperature to 0 to ensure a more accurate and deterministic behavior of the LLMs.

\subsection{Additional Experimental Details}
\label{appendix:experiment-details}

For the experiments we conducted in \S\ref{sec:exp}, the specific settings can be found in the training and configuration scripts we released.
All the experiments are performed using 1 - 4 NVIDIA A6000 GPUs with 48GB memory.
The experiments in low-resource settings take around 4 hours to converge, while the ones in high-resource settings take around 30 hours.

\noindent\textbf{Implementation Details of Contrastive Learning}
In practice, we observe that the magnitude of the log-probability in Eq.~\ref{eq:ctr-1} is highly dependent on the length of the candidate summaries.
Therefore, we introduce a modification to Eq.~\ref{eq:ctr-1} based on the length-normalized log-probability $\bar{p}_g$:
\begin{equation}
\label{eq:log-normalized}
\scalemath{0.9}{\bar{p}_g(S|D) = \frac{\sum_{i=1}^{l_S} \log p_g(s_i|S_{<i}, D)}{l_S},}
\end{equation}
and Eq.~\ref{eq:ctr-1} is changed to
\begin{equation}
\label{eq:ctr-2}
\begin{split}
    \mathcal{\hat{L}}_{ctr}(\theta) &= \sum_{S_i, S_j \in \mathcal{S}_c, i < j} \max(0,  \bar{p}_g(S_j|D; \theta) \\
    & - \bar{p}_g(S_i|D; \theta) + \frac{1}{\lambda}\log 2(j - i)),
\end{split} 
\end{equation}
where $\lambda$ is a scaling factor approximating the average summary length, which is set to the average length of the reference summaries.
As for the weight of contrastive loss ($\alpha$) in Eq.~\ref{eq:multi}, we performed a grid search to find the correct configuration, which is set to 100 for the low resource setting and 10 for the high resource setting.

\paragraph{Candidate Generation for Contrastive Learning}

The contrastive learning (Eq.~\ref{eq:ctr-2}) requires a list of candidate summaries.
To generate the summaries, we use the LLMs fine-tuned with MLE training and leverage diverse beam search as the sampling algorithm.
For training with GPTScore (\S\ref{subsec:gptscore}), we set 8 beam groups and 4 beams in each group, and pick one candidate from each group as the final candidate.
As for training with GPTRank (\S\ref{subsec:exp-gptrank}), we choose a larger search space with 32 beam groups, and pick 8 candidate outputs for the resulting 32 initial candidates by minimizing the similarity between them.
This is to ensure the diverse quality of candidate summaries used with GPTRank.
For the high resource training setting (\S\ref{subsec:llama2}), we follow a similar approach but use nucleus sampling~\cite{Holtzman2020The} instead of beam search for candidate generation to ensure score diversity from reference-based evaluation.

\subsection{Prompt Templates for GPTRank}
\label{appendix:gptrank}
In \S\ref{subsec:exp-gptrank}, we use the following prompt template for GPTRank with \textit{list-wise} comparison that is used for contrastive learning:

\begin{quote}
  You will be given a news article along with a list of summaries numbered as follows: 1. Summary 1, 2. Summary 2, and so on. 
Please evaluate and rank the summaries in descending order of their quality. First you will give an explanation of your ranking, then you will provide the ranking itself.
Please refer to the example below for the format of your response.

Example Response:

Explanation: ``Your explanation of the ranking''

Ranking: ``The ranking, e.g., 4, 2, 7, 3, 5, 6, 8, 1''

Here are the actual article and summaries:

Article:
\{\{Article\}\}

Summaries:

1. \{\{Summary 1\}\}

2. \{\{Summary 2\}\}

3. \{\{Summary 3\}\}

4. \{\{Summary 4\}\}

\end{quote}

For \textit{pairwise} comparison that is used for model evaluation, the prompt template is as follows:

\begin{quote}
    You will be given a news article along with two summaries.
Please compare the quality of these two summaries and pick the one that is better (there can be a tie).
First you will give an explanation of your decision then you will provide your decision in the format of 1 or 2 or tie.

Response format:

Explanation: ``Your explanation here''.

Decision: 1 or 2 or tie.

Here's the article:

\{\{Article\}\}

Summary 1:

\{\{Summary 1\}\}

Summary 2:

\{\{Summary 2\}\}
\end{quote}

\subsection{Analysis of Experiments with FLAN-T5}
\label{subsec:t5-analysis}

\begin{table}[t!]
\small
\centering
\begin{tabular}{lccc}
\toprule
\textbf{System} & \textbf{ROUGE-1}  & \textbf{ROUGE-2} & \textbf{Length}\\
\midrule
GPT-4       & 100.00  & 100.00  & 90.0   \\
\midrule
 BART.GPT4   & 63.22   & 44.70   & 91.8   \\
 BRIO.GPT4   & 58.65   & 37.57   & 92.8   \\
 T5.GPT4     & 62.99   & 44.31   & 93.9   \\
 T5BRIO.GPT4 & 58.44   & 36.69   & 108.4  \\
\bottomrule
\end{tabular}
\caption{Performance comparison of FLAN-T5 and BART as the fine-tuned backbone model on CNNDM.
\textbf{GPT-4} is the reference LLM.
\textbf{BART.GPT-4} and \textbf{T5.GPT-4} are fine-tuned with MLE training while \textbf{BRIO.GPT-4} and \textbf{T5BRIO.GPT-4} are fine-tuned with contrastive learning.
}
\label{tab:gptrank-t5} 
\end{table}

The reference-based evaluation results of the FLAN-T5 fine-tuning (\S\ref{subsec:comparative-study}) are reported in Table \ref{tab:gptrank-t5}.
We found that the FLAN-T5 checkpoint fine-tuned with contrastive learning, T5BRIO.GPT-4, tends to generate longer summaries.
We tried to control the summary length by adjusting the length penalty used during beam search, but found that the length difference was still present. 
On the other hand, we are able to control the summary length of BRIO.GPT-4.
We hypothesize this is because FLAN-T5 can learn the preference of LLM-based evaluation more efficiently, which exhibits a preference for longer outputs~\cite{rajani2023llm_labels}.
However, we note that the length preference is not the only factor affecting the LLM-based evaluation, since we only found a moderate Spearman's correlation (0.2366) between the summary length and the ranking of GPTRank.
Moreover, out of 20 summary pairs where the GPT-3.5 summary is longer than the T5BRIO.GPT-4 summary, T5BRIO.GPT-4 still wins 9 times as evaluated by GPTRank based on GPT-4.

\subsection{Experimental Details on XSum}
\label{subsec:xsum-appendix}

\begin{table}[t!]
\small
\centering
\begin{tabular}{lccc}
\toprule
\textbf{System} & \textbf{ROUGE-1}  & \textbf{ROUGE-2} & \textbf{Length}\\
\midrule
 GPT-4     & 100.00  & 100.00  & 42.8   \\
\midrule
 BART      & 31.90   & 12.35   & 21.8   \\
 BART.GPT4 & 56.45   & 35.82   & 42.6   \\
 BRIO.GPT4 & 57.08   & 36.55   & 42.9   \\
\bottomrule
\end{tabular}
\caption{Reference-based evaluation results on XSum.
\textbf{GPT-4} is the reference LLM.
\textbf{BART.GPT-4} is fine-tuned with MLE training while \textbf{BRIO.GPT-4} is fine-tuned with contrastive learning.
}
\label{tab:gptrank-xsum} 
\end{table}

The experimental setting on XSum (\S\ref{subsec:comparative-study}) is similar to the setting on CNNDM (\S\ref{subsec:setting}).
Specifically, at the warm-start stage we generate around 10K summaries using GPT-3.5 to fine-tune the BART checkpoint pre-trained on the original XSum dataset (\url{https://huggingface.co/facebook/bart-large-xsum}).
Then, we generate 1K summaries using GPT-4 and continue fine-tuning the checkpoint with MLE training, resulting in the checkpoint named BART.GPT-4.
As for contrastive learning, we use GPTRank with GPT-4 to generate 500 examples, and the checkpoint from the warm-start stage is fine-tuned to a new checkpoint, BRIO.GPT-4.
In Table~\ref{tab:gptrank-xsum}, we report the reference-based evaluation results.

\section{Human and Meta Evaluation Details}

\subsection{Definition of Summary Quality Aspects}
\label{subsec:aspect}
We adopt the definition of the different quality aspects in \S\ref{subsec:human-eval-collection} from the previous work~\citep{fabbri-etal-2021-summeval, gehrmann-etal-2021-gem, Gehrmann2022GEMv2MN} as follows:

\noindent (1) Salience: ``This rating measures how well the summary captures the key points of the news article. Consider whether all and only the important information are included in the summary.''

\noindent (2) Coherence: ``This rating measures whether the summary is presented in a clear, well-structured, logical, and meaningful way.''

\noindent (3) Overall Preference/Quality: ``This rating measures how much you like the summary.''

\subsection{Expert Evaluation Examples}
\label{subsec:annotation-examples}
We present expert-annotated examples (\S\ref{subsec:human-eval-results}) in Table~\ref{tab:expert-example}, with two main scenarios: (1) cases where the annotators unanimously favor LLM summaries; (2) cases where both LLM and smaller LM have good performance, resulting in different annotator preferences.
For those examples on which the annotators have different preferences for the \textit{overall} summary quality, we provide their explanations written \textit{after} the evaluation below, as a case study of the inherent subjectivity of summarization human evaluation.

\noindent \textbf{Example 3}

\noindent \textbf{Annotator 1}:
  {I selected the BRIO.GPT-4 summary because it conveys the same information as the GPT-3.5 summary more concisely. In the sentence about the nation being split on whether Charles should become king, it felt a little repetitive for the GPT-3.5 summary to use ``become king'' and ``ascend to the throne'' in the same sentence.}

\noindent \textbf{Annotator 2}:
    {The summaries are nearly identical. Both summaries capture almost the same level of important information. However, I prefer GPT-3.5's summary because it reiterates the fact that public opinion is expressed through a poll, which adds grounding and enhances objectivity to the statements.}

\noindent \textbf{Annotator 3}:
    {These two summaries essentially convey the same information and are almost equivalent in clarity and brevity. I chose the first summary because I personally preferred the way it started with (``A poll conducted by …'') which gives me the source of information that I value more.}

\noindent \textbf{Example 5}

\noindent \textbf{Annotator 1}:
   {I selected the GPT-3.5 summary because I found it slightly easier to follow along. The first two sentences both start with statements from Sheriff Hodgson, creating a clear structure and line of reasoning. The last sentence of the BRIO.GPT-4 summary ends with ``Hodgson said'', which makes sense but does not contextualize the statement until the very end of the summary.}

\noindent \textbf{Annotator 2}:
   {Both summaries are of good quality, making it a difficult decision for me. Despite its lower coherence and fluency, I lean towards preferring the summary generated by BRIO.GPT-4 due to its conciseness. The summary from GPT-3.5 includes additional details such as the mention of ``maximum-security Souza-Baranowski state prison'' and provides extra descriptions regarding Hernandez's charm, which I personally find redundant. }

\noindent \textbf{Annotator 3}:
   {Both summaries are of good quality, in terms of salience and coherence. The first one provides additional context regarding the final outcome of Aaron Hernandez's sentence, which I found to be more informative than the second summary.}

\noindent \textbf{Example 6}

\noindent \textbf{Annotator 1}:
    {I selected the GPT-3.5 summary because its first and last sentences were slightly more cohesive. The first sentence mentions that Nike has ``faced criticism'' and the last sentence mentions that Nike's vice president ``defended the decision'' in a statement -- a direct response to the criticism. On the other hand, the BRIO.GPT-4 summary starts by stating that Nike has ``defended their new kits'' but does not include any comments or defense from Nike.  }

\noindent \textbf{Annotator 2}:
    {I find it challenging to determine a clear winner between the two summaries as they both possess merits and weaknesses. The summary generated by GPT-3.5 mentions the key figure, Vice President Charlie Brooks, who defended the design of the kits, but it overlooks any feedback from the team. On the other hand, the summary generated by BRIO.GPT-4 fails to mention Charlie Brooks but includes the players' reactions, although it does so in a slightly redundant manner by quoting the midfielder Tobin Heath. In my opinion, the advantages and disadvantages of each summary are relatively balanced, leading me to consider them on equal footing.}

\noindent \textbf{Annotator 3}:
    {The second summary contains more balanced perspectives from both the critics and the national team itself. It also follows an organized structure from introducing the criticism to the reaction of the team. However, the first summary appears to be more straightforward and neutral, without individual responses and words such as ``proud'' which could create certain ambiguity for me. As such, I chose tie because they both match well with the purpose of a summary.}

\noindent \textbf{Example 7}

\noindent \textbf{Annotator 1}:
    {I selected the BRIO.GPT-4 summary because the last sentence provides specific, key information about Liana Barrientos' legal charges that are not mentioned in the GPT-3.5 summary, which provides important context for the case's current status.}

\noindent \textbf{Annotator 2}:
    {The quality of both summaries is high. GPT-3.5 mentions that all of Barrientos's marriages took place in New York State starting from 1999, which is a detail not mentioned in BRIO.GPT-4. While BRIO.GPT-4 does mention the crucial fact that some of Barrientos's partners could potentially pose threats to homeland security, I found the last sentence to be somewhat grammatically awkward. Therefore, I ultimately gave the edge to GPT-3.5.}

\noindent \textbf{Annotator 3}:
    {I prefer the second summary because it provides more essential details about the charges that Liana faces, including ``filing a false instrument'' and ``faces two counts of felony fraud charges''. Compared with the first summary, which essentially reiterates that Liana is a ``serial bride'', the second summary gives more emphasis to the legal aspect and the potential implications of her case.}

\subsection{LLM-based Evaluation Setting}
\label{subsec:llm-metric-setting}
In \S\ref{subsec:human-eval-results}, we compare the performance of different LLM-based evaluation methods.
Specifically, for G-Eval~\cite{liu-etal-2023-g} and GPTRank, we use different prompts for different quality aspects as we defined in Appendix \ref{subsec:aspect}.
The prompt templates we used for GPTRank are similar to the one shown in Appendix \ref{appendix:gptrank}, with specific quality aspect definitions.
As for G-Eval, the prompt is as follows (using the overall quality aspect as an example):
\begin{quote}
    You will be given one summary written for a news article.

Your task is to rate the overall quality of the summary with a score from 1 to 5, where 1 is the lowest and 5 is the highest.

Please make sure you read and understand these instructions carefully. Please keep this document open while reviewing, and refer to it as needed.

Evaluation Steps:

1. Read the news article carefully and identify the main topic and key points.

2. Read the summary and compare it to the news article. Check if the summary covers the main topic and key points of the news article, and if it presents them in a clear and logical order.

3. Assign a score for the summary quality on a scale of 1 to 5, where 1 is the lowest and 5 is the highest.

Input:

News Article:

\{\{Article\}\}

Summary:

\{\{Summary\}\}

Evaluation Form (scores ONLY):

- Overall Quality (1-5):
\end{quote}

For both GPTRank and G-Eval, we set the sampling temperature to 0 to enable more deterministic behaviors.
We note G-Eval tends to generate the same scores for different summaries as discussed in \citet{liu-etal-2023-g}, which likely results in its low agreement with human evaluation.

\begin{table*}[t]
    \scriptsize
    \centering
    \extrarowheight=\aboverulesep
    \addtolength{\extrarowheight}{\belowrulesep}
    \aboverulesep=0pt
    \belowrulesep=0pt
    \addtolength{\tabcolsep}{-2.0pt}
    \begin{tabular}{@{} c p{0.32\linewidth} p{0.32\linewidth} c c c}
\toprule
\textbf{ID} & \textbf{GPT-3.5} & \textbf{BRIO.GPT-4} & \textbf{Annotator 1} & \textbf{Annotator 2} & \textbf{Annotator 3}\\
\midrule
1 & A giant turnip weighing 33lbs and measuring 4ft long has been grown in China's Yunnan Province. The vegetable was grown naturally without the help of any fertilizers and has become a big attraction, with pictures of it going viral. This is not the first time such outsized vegetables have turned up in China, but usually they are grown by farmers using commercial fertilizers. & \cellcolor{gray!10} A giant turnip weighing 33lbs and measuring 4ft long has been grown in China's Yunnan Province. The monster vegetable, which weighs more than a three-year-old child, has been nicknamed the `fat little girl' in the village where it was grown and pictures of it have gone viral. The specimen was grown naturally without the help of any fertilizers and was so large it hangs over the flower bed. & GPT-3.5 & \cellcolor{gray!10} GPT-3.5 & GPT-3.5\\ 
 \midrule
2 & Wales midfielder Aaron Ramsey has warned his Arsenal team-mates that Wales could overtake England in the FIFA rankings. Wales are currently in 22nd place, their highest-ever position, and are unbeaten in Euro 2016 qualifying. Ramsey admits that there is already banter between himself and his English team-mates at Arsenal, and that Wales are catching up in the world rankings. & \cellcolor{gray!10} Arsenal midfielder Aaron Ramsey has warned his English team-mates to beware of Wales overtaking them in the FIFA rankings, as Wales climbed to their highest-ever position in football's world order in the April rankings to move within eight places of England. Chris Coleman's side are unbeaten in Euro 2016 qualifying and would be within touching distance of the finals in France should they beat Belgium in June, and Ramsey admits the banter with the likes of Theo Walcott, Jack Wilshere and Danny Welbeck is already flying on the Arsenal training ground. & GPT-3.5 & \cellcolor{gray!10} GPT-3.5 & GPT-3.5\\ 
 \midrule
3 & A poll conducted by the Daily Mail has revealed that the majority of the British public do not want Camilla to become queen if Prince Charles succeeds to the throne. The poll also showed that the nation is split on whether Charles should become king at all, with only 43\% believing he should ascend to the throne compared to an almost identical number saying he should stand aside in favour of his elder son, Prince William. William and Prince Harry are the most popular members of the Royal Family, closely followed by the Queen and the Duchess of Cambridge. & \cellcolor{gray!10} A majority of the British public do not want Camilla to become queen if Prince Charles succeeds to the throne, according to a poll for the Daily Mail. The nation is completely split on whether Charles should become king at all, with only 43\% believing he should, compared to an almost identical number saying he should stand aside in favour of his elder son Prince William. William and Prince Harry are the most popular members of the Royal Family, closely followed by the Queen and the Duchess of Cambridge. & BRIO.GPT-4 & \cellcolor{gray!10} GPT-3.5 & GPT-3.5\\ 
 \midrule
4 & A Russian warship and two bombers were intercepted by the UK's Royal Navy and RAF after they strayed close to UK territory, prompting suspicions that Russia was using access to a Nato war games exercise in Scotland to test UK reaction times. The incidents are the latest in a series of apparent tests by Moscow as tensions between Britain and Russia rise. Sir Michael Graydon, former head of the RAF, called for higher defence spending to defeat a ``resurgent'' Russia, saying that the Kremlin was using Cold War tactics to ``snoop'' on the exercise. & \cellcolor{gray!10} Russian warships with missiles, anti-aircraft guns and torpedoes entered the English Channel, forcing the UK to send a heavily-armed frigate to monitor the vessels. Two Russian long-range Bear bombers, capable of carrying nuclear missiles, hugged British airspace, prompting the RAF to deploy two Typhoon fighter jets, raising suspicions that Vladimir Putin was using access to the site to test UK reaction times. The incursions are the latest in a long line of apparent tests by Moscow as tensions between Britain and Russia rise. & GPT-3.5 & \cellcolor{gray!10} GPT-3.5 & GPT-3.5\\ 
 \midrule
5 & Bristol County Sheriff Thomas Hodgson has said that Aaron Hernandez, the former New England Patriots player who was sentenced to life for murder, is a master manipulator and will probably do fine in prison. Hodgson said Hernandez knows how to use his charm and manipulate better than anyone he has ever seen, adding that the former football star is generally affable and polite and would try to use those qualities to get what he wanted at the Bristol County House of Corrections. Hernandez will ultimately end up at the maximum-security Souza-Baranowski state prison. & \cellcolor{gray!10} Former New England Patriots player Aaron Hernandez has been sentenced to life in prison for the 2013 killing of Odin Lloyd, who was dating the sister of Hernandez's fiancee. Bristol County Sheriff Thomas Hodgson, who had Hernandez in custody for over 18 months, said that Hernandez is a master manipulator and will probably do fine in prison. Hernandez is generally affable and polite and would try to use those qualities to get what he wanted at the Bristol County House of Corrections, Hodgson said. & GPT-3.5 & \cellcolor{gray!10} BRIO.GPT-4 & GPT-3.5\\ 
 \midrule
6 & Nike has faced criticism for the new kits they have designed for the US women's national team, which do not feature the American flag. The kit is all white with hints of black and yellow, and the aqua colour cleats are a tribute to Canada, who are hosting the tournament. Nike's vice president of North America communications, Charlie Brooks, has defended the decision, stating that not all national teams have to represent the flag. & \cellcolor{gray!10} Nike has defended their new kits for the US women's national team following criticism that they do not represent the American flag. The latest design features an all white strip with hints of black and yellow, while the aqua colour cleats are a tribute to Canada who will be hosting the tournament. While patriotic supporters are up in arms, feedback from the team itself has been more positive, with midfielder Tobin Heath saying she is ``proud'' of the new designs. & GPT-3.5 & \cellcolor{gray!10} Tie & Tie\\ 
\midrule
7 & Liana Barrientos, an alleged ``serial bride'' accused of running a wife-for-hire immigration scheme, was arrested for evading a subway fare just after leaving court on Friday. Barrientos pleaded not guilty to charges that she married 10 men over 11 years and charged a fee for her ``services''. She has been accused of accepting money in at least one of the marriages and all of her marriages took place in New York state, allegedly starting in 1999. & \cellcolor{gray!10} Liana Barrientos, a woman accused of running a wife-for-hire immigration scheme, was arrested for evading the fare at a Bronx subway station after leaving court. She is accused of marrying 10 men over 11 years and charging a fee for her services, some of whom could pose a threat to American safety, according to investigators. She pleaded not guilty to two felony charges of filing a false instrument, involving marriage licences, and faces two counts of felony fraud charges. & BRIO.GPT-4 & \cellcolor{gray!10} GPT-3.5 & BRIO.GPT-4\\ 
\bottomrule
\end{tabular}
 \addtolength{\tabcolsep}{2.0pt}
\caption{Expert annotation examples of the pairwise comparison between GPT-3.5 and BRIO.GPT-4.
We show the three expert annotators' ratings regarding the \textit{overall} summary quality. 
}
\label{tab:expert-example}
\end{table*}

\end{document}